\pdfoutput=1

\documentclass[11pt]{article}

\usepackage[final]{acl}

\usepackage{times}
\usepackage{latexsym}

\usepackage[T1]{fontenc}

\usepackage[utf8]{inputenc}

\usepackage{microtype}

\usepackage{inconsolata}

\usepackage{graphicx}

\usepackage{times}
\usepackage{latexsym}
\usepackage{booktabs}
\usepackage{amsfonts}
\usepackage{nicefrac}
\usepackage{microtype}
\usepackage{todonotes}
\usepackage{listings}
\usepackage{algorithm}
\usepackage{algpseudocode}
\usepackage{soul}
\usepackage{tabularx}
\usepackage{pifont}
\usepackage{longtable}

\usepackage{xcolor}
\usepackage{amsmath}
\usepackage{hyperref}
\usepackage{cleveref}
\usepackage{comment}
\usepackage{tcolorbox}
\usepackage{nicematrix}

\usepackage{colortbl}
\usepackage{makecell}

\usepackage[normalem]{ulem}

\crefformat{section}{\S#2#1#3}
\crefformat{subsection}{\S#2#1#3}
\crefformat{subsubsection}{\S#2#1#3}
\crefrangeformat{section}{\S\S#3#1#4 to~#5#2#6}
\crefmultiformat{section}{\S\S#2#1#3}{ and~#2#1#3}{, #2#1#3}{ and~#2#1#3}

\definecolor{mypink1}{rgb}{0.858, 0.188, 0.478}

\definecolor{olive}{rgb}{0.39, 0.875, 0.12}

\newcommand{\eg}{{\it e.g.}}%
\newcommand{\ie}{{\it i.e.}}%

\newcommand{\ourbench}{TSG Bench}

\definecolor{yellow-green}{rgb}{0.3, 0.5, 0.0}

\definecolor{lightblue}{RGB}{224,236,247}
\definecolor{deepblue}{RGB}{9,46,107}
\definecolor{defaultcolor}{gray}{0.9}

\usepackage[T1]{fontenc}

\usepackage[utf8]{inputenc}

\usepackage{microtype}
\usepackage{algpseudocode,algorithm}

\usepackage{multirow}
\usepackage{graphicx}
\usepackage{pdfpages}
\usepackage{url}
\usepackage{epsfig}
\usepackage{adjustbox}
\usepackage{amsfonts}

\usepackage{amssymb}
\usepackage{booktabs}
\usepackage{comment}
\usepackage{arydshln}
\usepackage{caption}
\usepackage{microtype}
\usepackage{float}

\usepackage{xspace}

\usepackage{graphicx}
\usepackage{subcaption}

\title{LLM Meets Scene Graph: Can Large Language Models Understand and Generate Scene Graphs? A Benchmark and Empirical Study}

\renewcommand{\thefootnote}{\fnsymbol{footnote}}

\author{
    Dongil Yang\textsuperscript{\rm 1}~~~
    Minjin Kim\textsuperscript{\rm 1}~~~
    Sunghwan Kim\textsuperscript{\rm 1}~~~
    Beong-woo Kwak\textsuperscript{\rm 1}~~~ 
    \\
    \textbf{Minjun Park}\textsuperscript{\rm 1}~~~
    \textbf{Jinseok Hong}\textsuperscript{\rm 2}~~~
    \textbf{Woontack Woo}\textsuperscript{\rm 2, \rm 3}~~~
    \textbf{Jinyoung Yeo}\textsuperscript{\rm 1}\thanks{Corresponding author}\\
    Department of Artificial Intelligence, Yonsei University \textsuperscript{\rm 1}\\
    Graduate School of Metaverse, KAIST\textsuperscript{\rm 2}\\
    Graduate School of Culture Technology, KAIST\textsuperscript{\rm 3}\\
    \texttt{\{wingu,jinyeo\}@yonsei.ac.kr}
}

\begin{document}
\maketitle

\renewcommand{\thefootnote}{\arabic{footnote}}
\setcounter{footnote}{0}  

\begin{abstract}
The remarkable reasoning and generalization capabilities of Large Language Models (LLMs) have paved the way for their expanding applications in embodied AI, robotics, and other real-world tasks. To effectively support these applications, grounding in spatial and temporal understanding in multimodal environments is essential. To this end, recent works have leveraged \textit{scene graphs}, a structured representation that encodes entities, attributes, and their relationships in a scene. However, a comprehensive evaluation of LLMs' ability to utilize scene graphs remains limited.
In this work, we introduce \textbf{Text-Scene Graph (TSG) Bench}, a benchmark designed to systematically assess LLMs' ability to (1) understand scene graphs and (2) generate them from textual narratives.
With \ourbench{}, we evaluate 11 prominent LLMs and reveal that, while models perform well on scene graph understanding, they struggle with scene graph generation, particularly for complex narratives.
Our analysis indicates that these models fail to effectively decompose discrete scenes from a complex narrative, leading to a bottleneck when generating scene graphs.
These findings underscore the need for improved methodologies in scene graph generation and provide valuable insights for future research.
The demonstration of our benchmark is available at \href{https://tsg-bench.netlify.app/}{https://tsg-bench.netlify.app}.\footnote{Our code and evaluation data are publicly available at
\href{https://github.com/docworlds/tsg-bench}{https://github.com/docworlds/tsg-bench}.} 

\end{abstract}

\section{Introduction}
Large language models (LLMs) have demonstrated impressive progress in various text-based tasks, such as question-answering and content generation, showcasing strong reasoning and generation capabilities \citep{brown2020language,touvron2023llama}. Nevertheless, extending these abilities to multimodal environments is challenging, particularly when spatial and temporal reasoning about object relationships and physical interactions is required~\citep{yan2023inherent}.

\begin{figure}[t!]
    \centering
    \includegraphics[width=0.98\linewidth]{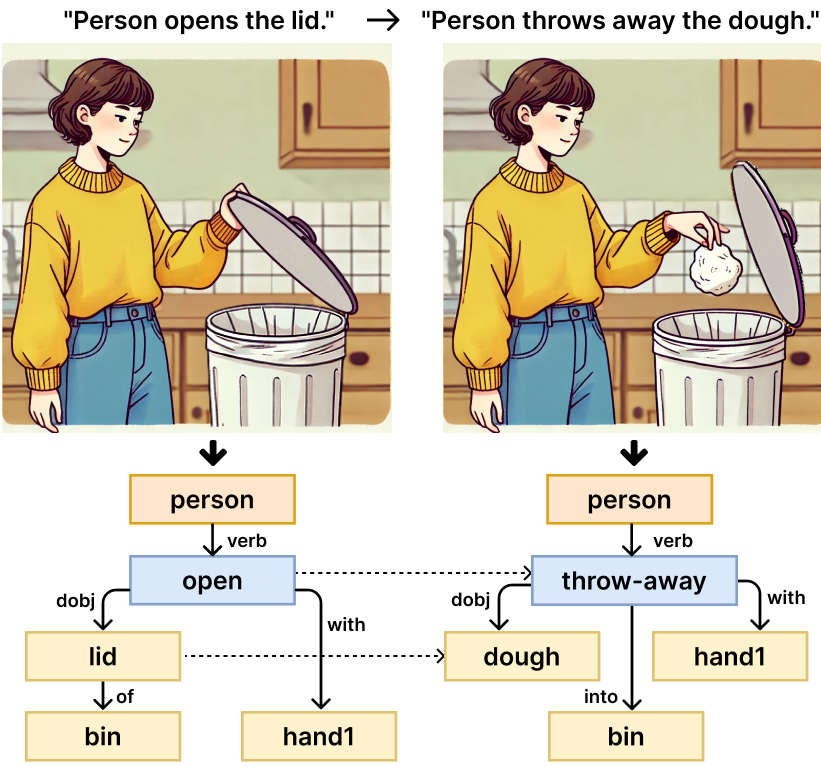}
    \caption{
        An illustration showing how a scene graph can represent 
        the objects and their relationships in a scene. 
        The illustration was created with the assistance 
        of DALLE-3.\protect\footnotemark
    }
    \label{fig:intro}
\end{figure}

\footnotetext{\url{https://openai.com/index/dall-e-3/}}

To address this issue, researchers have explored leveraging \textit{scene graphs} for LLMs \citep{chang2021comprehensive}. Scene graphs are structured representations that have been utilized in computer vision and embodied AI to capture key elements (\eg, objects, their attributes, and relationships) in complex multimodal environments \citep{ji2020action}. As illustrated in Figure~\ref{fig:intro}, by converting visual data into an interpretable representation, scene graphs enable LLMs to effectively understand spatial and semantic information, thereby allowing them to leverage their robust reasoning and generative capabilities in multimodal contexts. 

This integration paves the way for diverse applications ranging from dynamic scene interpretation to 3D environment modeling~\citep{Gao2023GraphDreamerC3,cong2023ssgvs,strader2024indoor,Zhang2024SceneLLMIL}.

Despite these advancements, a comprehensive evaluation of LLMs' ability to interpret and generate scene graphs remains limited\textemdash leaving open questions such as whether these models genuinely comprehend the underlying spatial and semantic structures. Bridging this gap is essential for developing systems that can perform reliable and structured reasoning across diverse domains. For example, LLMs often struggle to identify critical nodes or edges~\citep{huang2024can} and might misinterpret the triplets in complex situations, particularly when handling long contexts~\citep{kim2024llm4sgg}. 

We introduce \textbf{Text-Scene Graph Bench (\ourbench{})}, a benchmark designed to rigorously evaluate LLMs’ ability in scene graph understanding and generation. TSG Bench comprises long-text narratives that describe real-world scenarios alongside corresponding sequences of scene graphs representing an actor's interactions with objects. In every task, contextual grounding is ensured via a preceding scenario—delivered either in text or graph format—to mirror practical applications. For understanding tasks, we assess the models’ ability to interpret and reason over scene graphs. For generation tasks, we assess how accurately models can generate structured scene graphs given descriptive narratives and preceding contexts of varying complexity.

Through extensive experiments on eleven prominent LLMs using \ourbench{}, we make three key observations. (1) LLMs exhibit strong performance on scene graph understanding tasks. However, they significantly underperform on generation tasks, especially when faced with narratives that should be implicitly decomposed into multiple actions (\eg, implicit and repeated actions). (2) Advanced techniques, such as in-context learning and chain-of-thought (CoT) prompting, can facilitate the ability of highly capable LLMs to represent and reason over scene graphs. (3) LLMs can effectively refine errors in scene graphs when guided with error types. These findings highlight the need for improved methodologies in scene graph generation and provide critical insights for applications of LLMs in multimodal environments in future research.

\section{Related Work}

\paragraph{Scene graph representation.}
Scene graphs, which encode semantic information and relationships between objects in a scene, are widely used in various tasks across vision, language, and multimodal domains such as image captioning, 3D reconstruction, interactive games and robotics.~\citep{Johnson2015ImageRU, Yao2018ExploringVR, Chatterjee2021VisualSG, Armeni20193DSG, Ammanabrolu2021LearningKG, Ammanabrolu2020GraphCR, Rosinol20203DDS}.
Previous research has leveraged scene graphs as an inductive bias to facilitate explicit reasoning in visual question answering tasks~\citep{teney2017graph,Hildebrandt2020SceneGR} or to provide explanations based on such knowledge~\citep{Shi2018ExplainableAE}. Scene graphs can also serve as semantic maps of real-world environments for robots, enabling higher-level reasoning for tasks like planning and navigation~\citep{rana2023sayplan, Dai2023OptimalSG, yin2024sg}.

\paragraph{Scene graph datasets.}Such works have been facilitated by several benchmarks and datasets, most of which is built on image-scene graph pairs~\citep{Krishna2016VisualGC, ji2020action}. 
While FACTUAL~\citep{Li2023FACTUALAB} suggests a text-based benchmark for scene graph parsing, the dataset is focused on static scenes which restricts their applicability to dynamic, real-world scenarios. In contrast, our work extends to dynamic scenarios, where spatial, temporal relationship between scenes are actively involved. Ego-centric Action Scene Graphs (EASG) dataset~\citep{rodin2024action} explores the relationship between time-evolving actions and scene graphs in video contexts.
In contrast, our benchmark, \ourbench{}, is designed to evaluate the two distinct ability of LLMs -- reasoning and generation -- to link textual narratives and dynamic scene graphs to reflect richer context.

\paragraph{LLMs for scene graphs.}
Large Language Models, with their powerful reasoning and generation capabilities, have been employed for both understanding and generating scene graphs across various tasks. Recent works on robotics utilized LLMs to reason over scene graphs or formulate high-level plans for navigation within complex environments~\citep{yin2024sg,rana2023sayplan}. Beyond understanding, LLMs have also been applied to generate scene graphs by representing multiple objects and relations in a complex text and incorporating commonsense reasoning for 3D content generation tasks~\citep{Gao2023GraphDreamerC3,wei2024planner3d}. Despite these advancements, a comprehensive evaluation framework to systematically assess LLMs' performance on scene graph understanding and generation has remained underexplored.

\begin{figure*}[t!]
    \centering
    \includegraphics[width=0.98\linewidth]{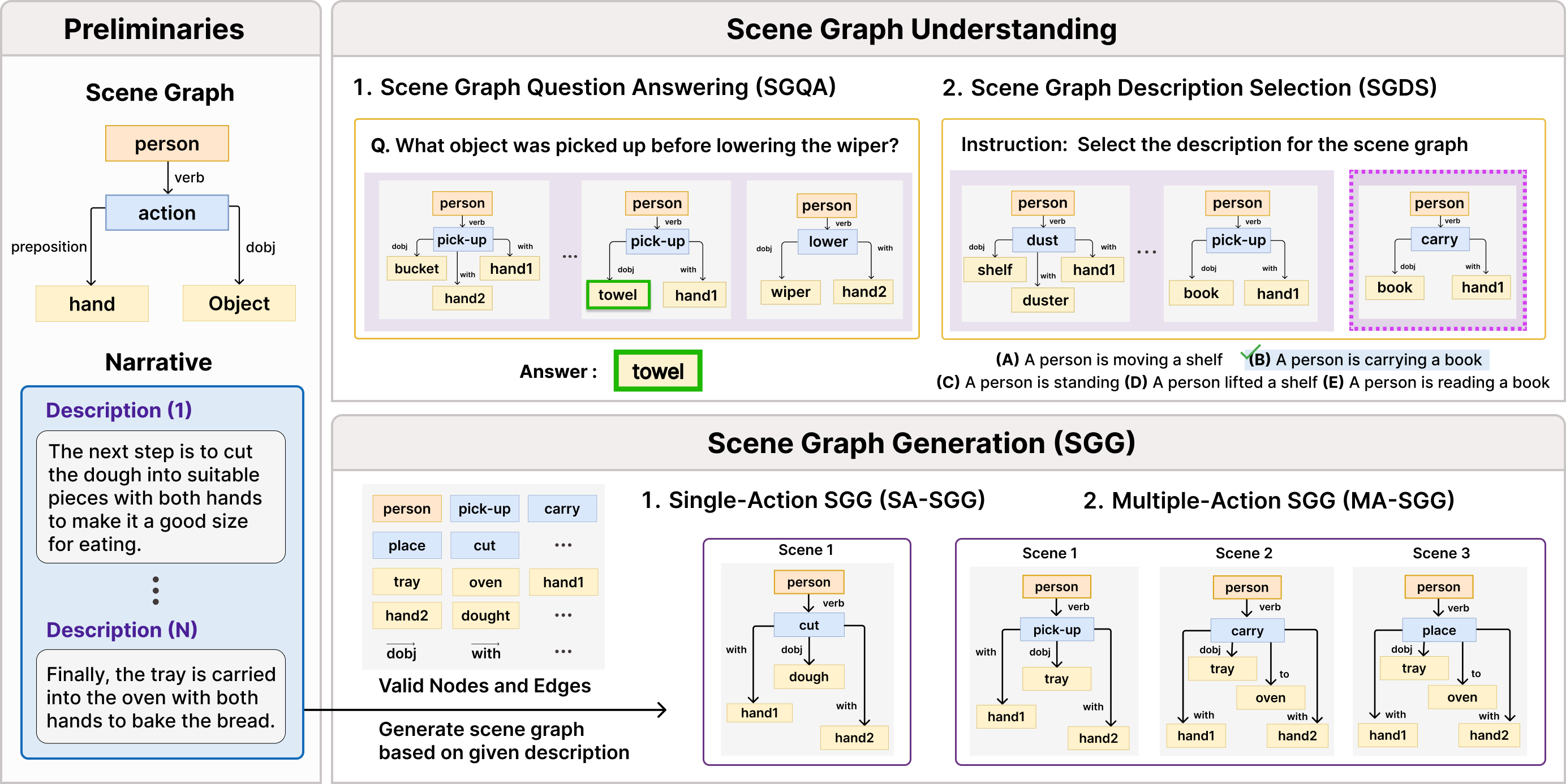}
    \caption{\textbf{Overview of \ourbench{}.} Scene graph question answering aims to answer a question by reasoning over scene graphs. Scene graph description selection is a multiple-choice task of selecting the correct description of a given scene graph. Single action scene graph generation focuses on generating a scene graph based on a description representing a single action. Multiple action scene graph generation aims to generate multiple discrete scene graphs of all actions represented in the description.}
    \label{fig:overview}
\end{figure*}

\section{\ourbench{}}~\label{bench}
In this section, we introduce \textbf{\ourbench{}}, a benchmark to evaluate LLMs' abilities to understand and generate scene graphs based on narratives. We begin by introducing the preliminaries of data representations, which form the basis of the benchmark.
Next, we detail the scene graph understanding and generation tasks, followed by an explanation of benchmark construction process. Finally, we provide the statistics of \ourbench{}.

\subsection{Preliminaries: Text and Scene Graph Representations}~\label{bench:composition}
We first construct a database of narratives and scene graphs representing sequential scenes of real-world scenarios, and configure understanding and generation tasks from it. The narrative is composed of multiple coherent natural language descriptions, denoted as $D=(d_1, \dots, d_n)$ where $n$ represents the number of descriptions in each scenario.
For each description $d_{i}$, a set of scene graphs $G_i = (G_{i1}, \dots, G_{ik})$ is aligned, where each scene graph represents an action, following the action-centric scene graph representation proposed in a previous work~\citep{rodin2024action}. In the case of our database, the number of scene graphs $k$ for a description range from 1 to 8, depending on the complexity of $d_i$. For example, when a description of \textit{shaking} an object is given, $k$=2 because it can be decomposed into \textit{holding} and \textit{rapidly moving} the object. For each $j$ such that $1 \leq j \leq k$, the individual scene graph is denoted as $G_{ij} = (V_{ij}, E_{ij})$, where $V_{ij}$ and $E_{ij}$ represent the nodes and edges of the scene graph, respectively. More specifically, the relationships between nodes and edges are expressed as a set of triplets in the form of (source node, edge, target node).

\paragraph{Nodes and edges.}
A node belongs to one of four categories \{\texttt{person}, \texttt{action}, \texttt{object}, \texttt{hand}\}, where \texttt{person} represents an actor, \texttt{action} denotes the actor's action in the scene, \texttt{object} refers to an item or a location, and \texttt{hand} corresponds to either \texttt{hand1} or \texttt{hand2} of the actor. Similar to previous works~\citep{rodin2024action,grauman2022ego4d}, we use \texttt{hand} nodes to track activities involving one or both hands. Specifically, once an item is held, it occupies \(\texttt{hand1}\); if a second item is acquired, \(\texttt{hand2}\) is assigned. Releasing all items resets the next grasp to \(\texttt{hand1}\). An edge belongs to one of three categories \{\texttt{verb}, \texttt{dobj}, \texttt{preposition}\}, where \texttt{verb} connects a \texttt{person} node to an \texttt{action} node, \texttt{dobj} links an \texttt{action} node to an \texttt{object} node only if it is the direct object of the action, and \texttt{preposition} connects any pair for representing the spatial relationship or other contextual dependencies between them. To ensure consistency and standardize elements (\ie, nodes and edges) in the graph, we define $L$ as a predefined collection of valid node and edge values for each scenario. Consequently, any additional modifiers or redundant expressions in descriptions $D$ are omitted to maintain a concise representation.

\subsection{Scene Graph Understanding}~\label{bench:understanding}

\paragraph{Scene graph question answering (SGQA).}
This task involves reasoning over scene graphs to answer a given question. Formally, given a question and scene graphs $G=(V, E)$, the model must predict an answer, which corresponds to an element in $V$. As shown in Figure~\ref{fig:overview}, questions in our benchmark require logically or temporally connecting a sequence of actions or object state changes, which can be solved by hopping across multiple triplets.

\paragraph{Scene graph description selection (SGDS).}
The goal of this task is to accurately interpret a scene graph within a given context and identify the correct description among distractors. We formulate SGDS as a multiple-choice question problem, consisting of the graph-based context $C^g_i=(G_1, \dots, G_{i-1})$, a scene graph $G_{i}$, and five candidate descriptions, with one correct answer included. The model should be able to track nodes and edges from $C^g_i$ and ensure that all elements in $G_i$ are accurately represented. For SGDS, we use scene graphs representing a single action.

\subsection{Scene Graph Generation}~\label{bench:generation}
Scene graph generation tasks aim to generate triplets of scene graphs corresponding to a given description within a context. All valid elements are predefined as $L$ for each scenario, and the tasks require models to identify and parse semantically similar elements from the given description to construct triplets. Each task is illustrated in Figure~\ref{fig:overview}.

\paragraph{Single action scene graph generation (SA-SGG).}
SA-SGG is a task of generating a scene graph for a description that involves a single action, within a scenario. Formally, the task is to generate triplets of $G_i=(V_i, E_i)$, given the description context $C^d_i=(d_1, \dots, d_{i-1})$, a description $d_i$, and valid nodes and edges of the scenario, denoted as $V$ and $E$, respectively.

\paragraph{Multiple action scene graph generation (MA-SGG).}
MA-SGG aims to generate scene graphs by decomposing actions when given complex descriptions that involve multiple actions. The task formulation is identical to that of SA-SGG, except that an additional clue indicating the number of actions is provided, and the complexity of $d_i$ is greater than 1. This makes MA-SGG more challenging than SA-SGG because the amount of information to process, especially to generate, is larger, and target actions may be implicit in the description. Also, although the number of actions is given, the task still requires the ability to accurately decompose, identify, and order valid actions from the description.

\subsection{Dataset Construction}~\label{bench:construction}
We derive \ourbench{} from the Ego-centric Action Scene Graphs (EASG) dataset~\citep{rodin2024action}, which represents temporally evolving actions in video contexts as scene graphs. We use the original scene graphs to build our own narratives and corresponding scene graphs through multiple rounds of a human-in-the-loop process involving three trained annotators.
First, we prompt an LLM to generate a sentence from each scene graph from the EASG dataset and remove redundant sentences based on context. After human workers review the logical flow and naturalness, we prompt the LLM again to generate a scene graph for each sentence. As they often fail to generate complete scene graphs, human workers meticulously inspect and refine the graph elements one by one. Then, we prompt the model to paraphrase sentences to increase lexical diversity and to combine coherent sentences, enhancing overall complexity. As a result, we collect 120 scenarios of 2,041 descriptions and 4,289 scene graphs.

\paragraph{Task-related data.}
We additionally construct data for our understanding tasks through a similar collaboration process.
For SGQA, we provide the LLM with the entire scenario narrative and prompt it to generate five questions about identifying a node that had undergone spatial and temporal transitions. Human workers then verify the validity of these questions. For SGDS, we control the difficulty of the distractors by perturbing the answer description in two ways. We put random distractors from unrelated scenarios for the half, and put distractors with LLM-perturbed nodes and edges from the answer for the other half of the problems. More detailed descriptions of our dataset construction process are in Appendix~\ref{sec:appendix_data_construction}.

\begin{table}[t!]
    \centering
    \normalsize
    \setlength{\tabcolsep}{8pt}
    \begin{tabular}{l c}
    \toprule
         \textbf{Statistics} & \textbf{Counts} \\ 
         \midrule
         \textbf{Benchmark Statistics} \\
         \# of Domains & 18 \\
         \# of Scenarios & 120 \\
         \# of Descriptions & 2,041 \\
         \# of Scene graphs & 4,289 \\
          \quad -- \ avg. \# nodes & 4.81 \\
         \quad -- \ avg. \# edges & 3.45 \\
         \# of Nodes & 14,905 \\
         \# of Edges & 11,820 \\
         \midrule
         \textbf{Task Statistics} \\
         \# data for SGQA & 500 \\
         \# data for SGDS & 250 \\
         \# data for SA-SGG & 1,188 \\
         \# data for MA-SGG & 853 \\
         \quad -- \ avg. \# scene graphs & 3.64 \\

         
    \bottomrule
    \end{tabular}
    \caption{Statistics of \ourbench{} and each task.}
    \label{tab:statistics}
    \vspace{-0.1in}
\end{table}

\begin{figure}[ht!]
    \centering
    \includegraphics[width=0.8\linewidth]{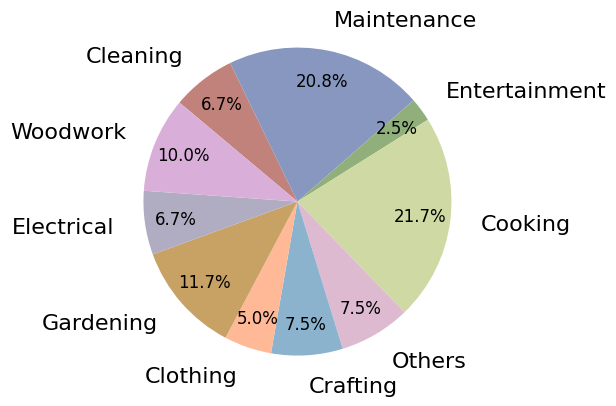}
    \caption{Domain distribution across scenarios.}
    \label{fig:domain}
\end{figure}
\subsection{Benchmark Statistics}~\label{bench:statistics}
Table~\ref{tab:statistics} summarizes the statistics of \ourbench{} and each task. Our benchmark provides 120 real-world scenarios, covering a wide range of domains, including maintenance, cooking, and gardening, as illustrated in Figure~\ref{fig:domain}. \ourbench{} contains 2,041 descriptions and 4,298 scene graphs, and 14,905 and 11,820 different nodes and edges, respectively, covering a wide semantic space. As we focus on action-centric scenarios, common action nodes include \emph{pick-up}, \emph{place}, \emph{hold}, and \emph{release}, and the most common preposition edge is \emph{with}, involving hand movements.

For understanding tasks, we construct 500 datapoints for SGQA, and 250 for SGDS. For generation tasks, there are 1,188 data samples for SA-SGG and 853 for MA-SGG. For MA-SGG, the average number of scene graphs to generate, or the complexity $k$, is approximately 3.64.

\begin{table*}[ht]
\centering
\small
\renewcommand{\arraystretch}{1}
\resizebox{\textwidth}{!}{%
\begin{tabular}{l c c c c c c c c}
\toprule
\textbf{Model}
& \textbf{SGDS} & \textbf{SGQA}
& \multicolumn{3}{c}{\textbf{SA-SGG}} 
& \multicolumn{3}{c}{\textbf{MA-SGG}} \\
\cmidrule(lr){2-2}\cmidrule(lr){3-3}\cmidrule(lr){4-6}\cmidrule(lr){7-9}
& \textbf{Accuracy} & \textbf{EM}
& \textbf{Precision} & \textbf{Recall} & \textbf{F1}
& \textbf{Precision} & \textbf{Recall} & \textbf{F1} \\
\midrule
Human
& 98.33 & 88.00
& 85.22 & 81.00 & 82.50
& 78.80 & 72.90 & 75.60 \\
\midrule
\multicolumn{9}{c}{\emph{Proprietary models}} \\
\midrule
GPT-4o
& 96.40 & 84.80
& 65.84 & 55.04 & 59.23
& 48.88 & 40.84 & 43.99 \\

GPT-4o-mini
& 96.80 & 76.60
& 20.00 & 21.50 & 19.90
& 23.06 & 18.32 & 20.07 \\

Claude-3.5-Sonnet
& \textbf{98.40} & \textbf{90.60}
& \textbf{69.75} & \textbf{69.33} & \textbf{68.43}
& \textbf{60.77} & \textbf{57.91} & \textbf{58.80} \\

Claude-3.5-Haiku
& 97.20 & 82.00
& 38.31 & 36.82 & 36.77
& 27.00 & 23.97 & 24.95 \\
\midrule
\multicolumn{9}{c}{\emph{Open source models}} \\
\midrule
LLaMA-3.3-70B
& \textbf{97.60} & \textbf{84.60}
& 31.52 & 38.90 & 33.37
& 32.43 & 26.58 & 28.92 \\

Qwen-2.5-72B
& 96.80 & 81.40
& 57.96 & 53.22 & 54.42
& 42.64 & 33.29 & 36.78 \\

DeepSeek-V3
& 96.40 & 79.60
& 55.79 & 55.11 & 54.45
& \textbf{43.67} & \textbf{36.66} & \textbf{39.34} \\

Mixtral-8x22B
& 96.00 & 73.00
& 31.03 & 32.79 & 30.84
& 21.05 & 19.54 & 19.75 \\

Mistral-large
& 96.40 & 82.40
& \textbf{63.18} & \textbf{55.37} & \textbf{58.15}
& 40.17 & 32.12 & 35.13 \\

\hdashline
Qwen-2.5-7B
& 93.60 & 73.40
& 9.58 & 9.95 & 9.39
& 6.61 & 6.77 & 6.34 \\

Mistral-7B
& 90.14 & 58.20
& 13.60 & 14.64 & 13.14
& 13.86 & 10.57 & 11.67 \\
\bottomrule
\end{tabular}
}
\caption{Main results for scene graph understanding tasks (SGDS, SGQA) and scene graph generation tasks (SA-SGG, MA-SGG). The full prompts are listed in Appendix \ref{appendix:eval_prompts}.}
\label{tab:main_sg2t_no_hard}
\end{table*}

\section{Experiments}

\subsection{Setup}
We conduct our experiments on eleven highly capable LLMs to provide a comprehensive assessment of current LLMs. (1) For proprietary models, we choose GPT-4o, GPT-4o-mini~\citep{openai2024gpt4o}, Claude-3.5-Sonnet, and Claude-3.5-Haiku~\citep{anthropic2024claude}; (2) For open-source models, we select LLaMA-3.3-70B~\citep{llama3_3}, Qwen-2.5-72B, Qwen-2.5-7B~\citep{qwen2024}, DeepSeek-V3~\citep{deepseekv3}, Mixtral-8x22B~\citep{mixtral8x22b}, Mistral-large, and Mistral-7B~\citep{mistrallarge}. Additionally, we also provide human performance on a subset of 30 examples to facilitate the interpretation of the results. Experiment details are in Appendix~\ref{appendix:experiment_details}.

\subsection{Evaluation Protocols}
We evaluate LLMs on their capabilities in scene graph understanding and generation by prompting them in a zero-shot fashion, using prompts listed in Appendix~\ref{appendix:eval_prompts}. We use different metrics for each task. We assess SGQA using Exact Match (EM), which requires the model's generation to match the reference element exactly. For SGQA, we instruct LLMs to generate a single letter representing the predicted candidate and evaluate it using accuracy. Scene graph generation tasks are assessed with precision, recall, and macro F1 score. For MA-SGG, where one description yields multiple scene graphs, evaluation is conducted separately for each generated graph.

\subsection{Main Results} \label{main_results}
We compare the scene graph understanding (SGDS, SGQA) and generation (SA-SGG, MA-SGG) performance of different models on \ourbench{} in Table \ref{tab:main_sg2t_no_hard}.

\paragraph{Scene graph understanding.}
Most models exhibit strong performance in scene graph understanding tasks, particularly in the SGDS task. Among the models, Claude-3.5-Sonnet shows relatively strong performance\textemdash 98.40 in SGDS and 90.60 in QA. Even though open source models such as LLaMA-3.3-70B achieve competitive performance (97.60 in SGDS, 84.60 in SGQA), no model consistently outperforms strong proprietary models such as Claude-3.5-Sonnet and GPT-4o. For the relatively smaller open-source models (Qwen-2.5B-7B, Mistral-7B), while the models demonstrate competitive performance in SGDS task (93.60 and 90.14 respectively), the performance still falls short in the SGQA task compared to other models with a larger number of parameters\textemdash 73.40 for Qwen-2.5-7B and 58.20 for Mistral-7B in EM.

\paragraph{Scene graph generation.}
In contrast to scene graph understanding tasks, scene graph generation remains challenging for LLMs. Similar to the trend in understanding tasks, Claude-3.5-Sonnet achieves the highest scores among the models -- F1 of 68.43 (SA-SGG) and 58.80 (MA-SGG). However, a notable difference is that Claude-3.5-Sonnet is no longer as good as a human in the scene graph generation task, scoring lower compared to human performance (82.50, 68.43). The performance gap between LLMs and humans is evident in generating multiple coherent scene graphs (MA-SGG). For example, the performance of Claude-3.5-Haiku, which is one of the most competitive models in SGQA, decreases by 58.8\% in MA-SGG. Many models also show higher precision than recall in this setting (\eg, Mistral-large: 40.17 vs. 32.12), indicating incomplete coverage of sub-scenes. This shortfall arises from the need to split a single description into multiple scenes and construct separate graphs for each, increasing structural complexity and reducing recall.
We provide deeper analysis on generation task by disentangling the task into three distinct subtasks in Section~\ref{analysis:multiple_generation_disentangled}. Open-source models perform worse, with Qwen-2.5-72B (66.15, 43.73) and Mistral-large (69.76, 37.76) being the strongest but still far behind proprietary models. Smaller models like Qwen-2.5-7B and Mistral-7B fail severely in MA-SGG ($< 12$ in F1).

\begin{figure*}[t!]
    \centering
    \includegraphics[width=0.98\linewidth]{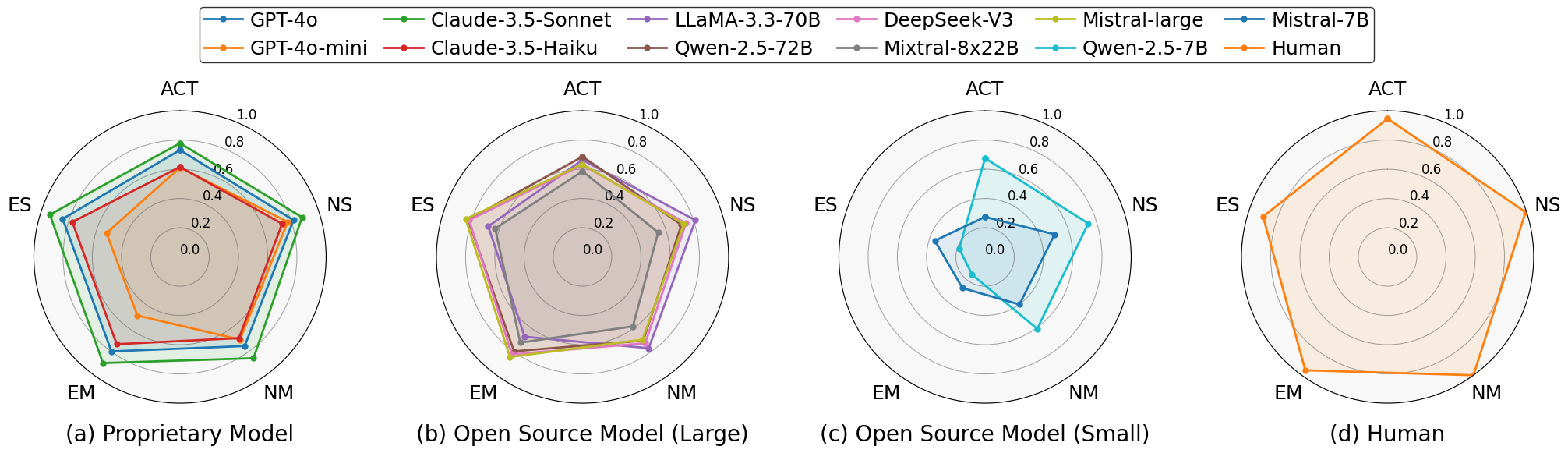}
    \caption{F1-score results on decomposed scene graph generation tasks, distinguishing between single-action and multiple-action settings. ES (Edge Single) and NS (Node Single) evaluate edge and node generation performance in SA-SGG, respectively. EM (Edge Multiple) and NM (Node Multiple) assess edge and node generation in MA-SGG. ACT (Action) measures the model’s performance in action decomposition in MA-SGG.}
    \label{fig:subtask}
\end{figure*}

\section{Analysis}

To provide further insight on leveraging LLM for scene graph tasks, we explore the following four research questions.

\subsection{Which Challenges Arise When LLMs Generate Scene Graphs?} \label{analysis:multiple_generation_disentangled}

The main results in Table~\ref{tab:main_sg2t_no_hard} indicate that language models struggle in generation tasks compared to understanding tasks. Hence, we aim to discover the underlying challenges by configuring three subtasks of scene graph generation: node generation, edge generation, and action decomposition. For a clearer analysis, we exclude action nodes in the reference for node generation subtask. Note that, both single and multiple action scene graph generation tasks, node and edge generation abilities are essential, and in multiple action scene graph generation task, action decomposition ability is additionally required.

Since a scene graph is represented with triplets of nodes and edges, we ablate the effect of each by conditioning on the other to generate either one. Following the notations described in Section~\ref{bench}, node generation task is formulated as $(C^d_i, d_i, L, E_i) \rightarrow V_i-\{action\}$. The edge generation task is formulated as $(C^d_i, d_i, L, V_i) \rightarrow G_i$, as edges are designed to represent the relationships between two nodes. Lastly, action decomposition aims to generate all actions in a correct sequence, given $C^d_i, d_i,$ and $L$.

According to Figure~\ref{fig:subtask}, we find that large models tend to generate edges better than nodes. On the other hand, small language models perform poorly on edge generation. Furthermore, in the MA-SGG, most models lack in decomposing actions correctly in complex descriptions.

\begin{figure}[t]
    \centering
    \includegraphics[width=\columnwidth]{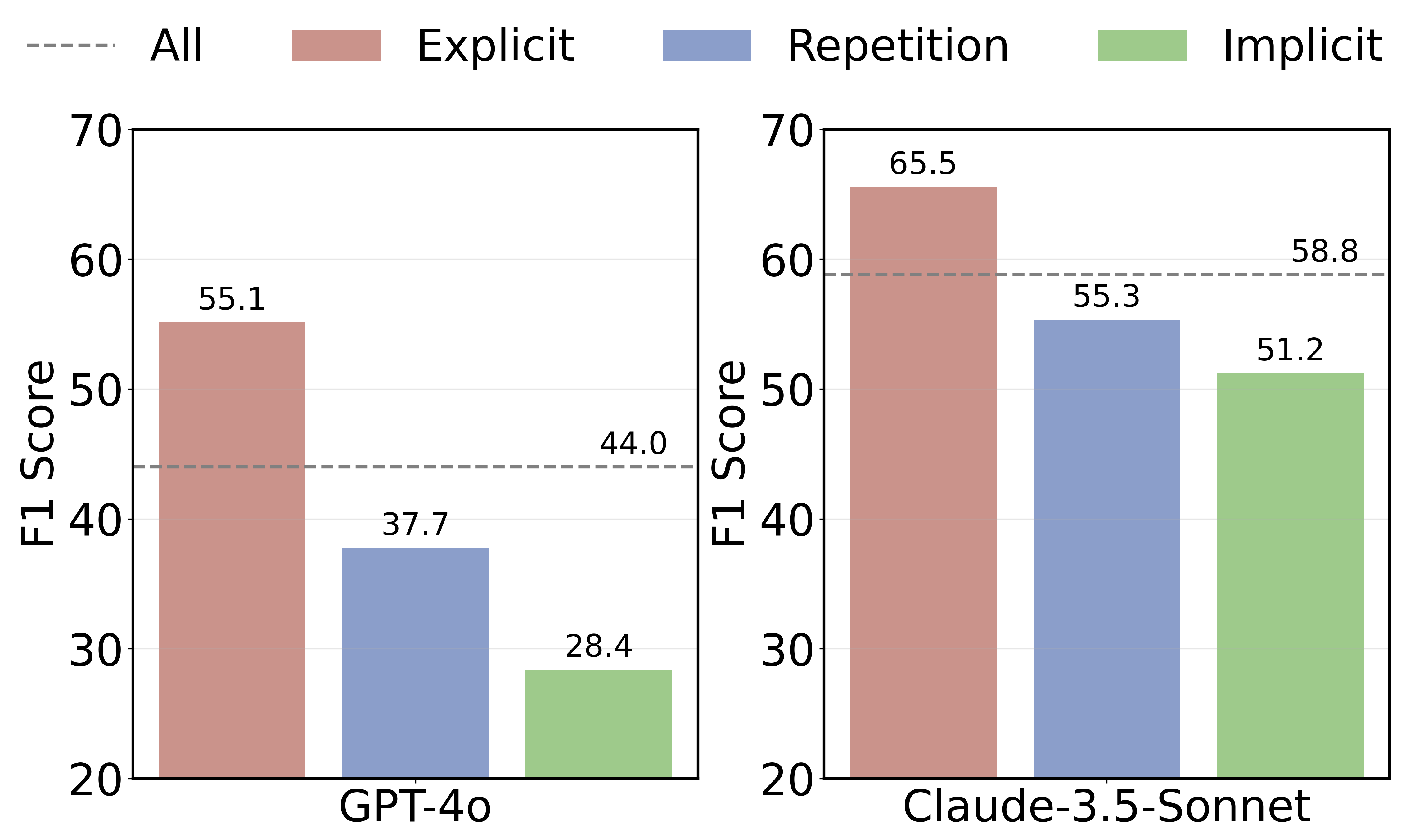}
    \caption{The results of the comparative evaluation under four conditions—Explicit, Implicit, Repetition, and All. The “All” condition comprises the entire dataset, while the other three focus on subsets featuring explicit actions, implicit actions, or repeated actions.}
    \label{fig:implicit}
\end{figure}

\begin{table}[ht]
\centering
\small
\renewcommand{\arraystretch}{1}
\begin{tabular}{@{} p{0.95\linewidth} @{}}
\toprule
\textbf{[Context]} \\
 ... The timber was carefully dehydrated to ensure it was ready for the next stage of the woodworking project. \\
\noalign{\vskip 5pt}
\hdashline
\noalign{\vskip 5pt}
\textbf{[Description]} \\
The timber was \textbf{clean}ed using a fabric.\\
\midrule
\textbf{[Nodes in the Reference Graph]} \\
$V_1$: person, \textbf{pick-up}, cloth, ...   \\
$V_2$: person, \textbf{wipe}, wood, ...  \\
\midrule
\textbf{[Nodes in the Generated Graph]} \\
$V_1$: person, \textbf{wipe}, cloth, ...  \\
$V_2$: person, \textbf{hold}, wood, ...  \\
\bottomrule
\end{tabular}
\caption{A failure case of action decomposition, which involves two consecutive actions: picking up a cloth and wiping the wood with it. \( V_j \) indicates the list of nodes in the \( j \)-th graph of the description.}
\label{tab:tsg_bench_stats}
\end{table}

\subsection{Which Action Do LLMs Struggle to Generate?}
We hypothesize that LLMs may struggle to represent structured graphs when descriptions contain implicit actions. To examine this, we curated a subset of descriptions in which referenced actions are not explicitly stated in any variant form. We then conducted MA-SGG experiment using GPT-4o and Claude-3.5-Sonnet. As shown in Figure~\ref{fig:implicit}, both models perform worse when implicit actions are present, while they perform better when generating explicit actions.

Through manual analysis on failure cases, we discovered another challenging type of actions, repetitive actions (\eg, \textit{sweep three times}). We curated another subset of descriptions containing repetitive actions by filtering for words such as \textit{times, repeat, and repeated}. As shown in Figure~\ref{fig:implicit}, the models struggle with generating the same action multiple times. Consistent with previous studies~\citep{brown2020language}, our results suggest that LLMs lack a strong numerical sense, which humans find trivial. Examples are shown in Appendix~\ref{appendix:repetition_case}.

\begin{table}[h]
\centering
\small
\resizebox{\columnwidth}{!}{%
\begin{tabular}{lcccc}
\toprule
\textbf{Method}
& \(\textbf{SGDS}\)
& \(\textbf{QA}\)
& \(\textbf{SA-SGG}\)
& \(\textbf{MA-SGG}\) \\
\midrule
\textbf{GPT-4o}          
& 96.40 & 84.80 & 59.23 & 43.99 \\
\quad + CoT              
& 96.80 & \textbf{90.00} & \textbf{67.13} & 44.79 \\
\quad + 10-shot          
& \textbf{99.20} & 84.40 & 65.78 & \textbf{57.40} \\
\midrule
\textbf{Claude-3.5-Sonnet}
& 98.40 & 90.60 & 68.43 & 58.80 \\
\quad + CoT              
& 98.00 & \textbf{94.00} & 69.57 & 64.36 \\
\quad + 10-shot          
& \textbf{98.80} & 92.00 & \textbf{75.29} & \textbf{71.75} \\
\midrule
\textbf{Qwen-2.5-72B}    
& 96.80 & 81.40 & 54.42 & 36.78 \\
\quad + CoT              
& \textbf{97.60} & \textbf{88.00} & 53.43 & 31.33 \\
\quad + 10-shot          
& \textbf{97.60} & 84.60 & \textbf{67.87} & \textbf{53.47} \\
\midrule
\textbf{Mistral-large}   
& 96.40 & 82.40 & 58.15 & 35.13 \\
\quad + CoT              
& 95.20 & \textbf{96.00} & 62.45 & 32.39 \\
\quad + 10-shot          
& \textbf{98.80} & 85.40 & \textbf{66.99} & \textbf{48.10} \\
\midrule
\textbf{Qwen-2.5-7B}     
& 93.60 & \textbf{73.40} & 9.39  & 6.34 \\
\quad + CoT              
& 94.00 & 72.00 & 11.56 & 3.88 \\
\quad + 10-shot          
& \textbf{95.60} & 73.20 & \textbf{39.96} & \textbf{37.25} \\
\midrule
\textbf{Mistral-7B}      
& 90.14 & 58.20 & 13.14 & 11.67 \\
\quad + CoT              
& 94.00 & \textbf{68.00} & 10.97 & 6.32 \\
\quad + 10-shot          
& \textbf{94.80} & 66.20 & \textbf{37.97} & \textbf{33.74} \\
\bottomrule
\end{tabular}
}
\caption{The results of each method are evaluated using the same setup as the main results. Rows listing only the model name correspond to the vanilla (zero-shot) setting. The detailed prompts for both CoT and few-shot approaches are provided in the Appendix \ref{appendix:CoT_prompts} and \ref{appendix:few_shot_prompts}.}
\label{tab:evaluation_cot}
\end{table}

\subsection{Do Advanced Prompting Methods Elicit Better Performance?}
We further investigate whether advanced prompting methods for improving LLM capabilities can benefit the models in Table \ref{tab:evaluation_cot}. To assess the effectiveness of popular LLM prompting techniques, we conduct experiments with Chain-of-Thought (CoT) prompting and 10-shot in-context learning (ICL). The results indicate that 10-shot ICL generally improved performance across tasks, particularly for SGDS, SA-SGG, and MA-SGG. These tasks require a deeper understanding and structured representation of scene graphs, making ICL an effective approach. In contrast, CoT prompting proved beneficial for reasoning-intensive tasks such as SGQA. However, the performance gains varied across models. For instance, Qwen-2.5-7B, which already exhibited relatively high initial performance, showed minimal improvement with CoT, potentially due to longer prompts introducing confusion. On the other hand, Mistral-7B, which initially demonstrated lower performance, benefited significantly from CoT, suggesting that reasoning augmentation can compensate for weaker baseline capabilities.

\subsection{Can LLMs Refine Errors in Scene Graphs?}

We assess whether LLMs have the ability to refine an incorrect scene graph. To conduct a controlled analysis, we curated 5,940 data samples with different types of errors in the scene graph: \emph{Redundant} (extra triplet), \emph{Missing} (omitted triplet), \emph{Mismatched} (perturbed element), and \emph{Reversed} (inverted directions). Given the context, the description, and the erroneous graph, we prompt GPT-4o and Claude-3.5-Sonnet to refine the graph to align with the meaning of $d$. As shown in Table~\ref{tab:refinement}, while both models generally underperform, it stands out for the \emph{Mismatched} type. These findings highlight the importance of precise interpretation and correction of subtle errors within scene graphs.

\begin{table}[t!]
\centering
\renewcommand{\arraystretch}{1}
\resizebox{\columnwidth}{!}{%
\begin{tabular}{lrrrr}
\toprule
\multirow{2}{*}{\textbf{Error Type}} 
   & \multicolumn{2}{c}{\textbf{\textit{w/o} Error Type}} 
   & \multicolumn{2}{c}{\textbf{\textit{w/} Error Type}} \\
\cmidrule(lr){2-3}\cmidrule(lr){4-5}
& \textbf{GPT-4o} & \textbf{Claude} 
& \textbf{GPT-4o} & \textbf{Claude} \\
\midrule
Overall      & 40.04 & 60.03 & 64.80 & 88.28 \\
Redundant    & 64.38 & 70.02 & 73.29 & 93.06 \\
Missing      & 46.67 & 82.25 & 58.42 & 80.89 \\
Mismatched   & 10.81 & 38.29 & 58.92 & 81.94 \\
Reversed     & 44.24 & 49.55 & 68.55 & 97.22 \\
\bottomrule
\end{tabular}%
}
\caption{
  Refinement results with and without error‐type awareness, evaluated with the F1-score. 
  “\textit{w/o} Error Type” denotes refinement without error type, and 
  “\textit{w/} Error Type” denotes refinement with error type.
  “Claude” refers to the Claude‐3.5‐Sonnet model.
}
\label{tab:refinement}
\end{table}

We further investigate whether the models’ refinement challenges arise from insufficiently detecting error types or from difficulties in correcting errors they have already identified. To address this, we evaluate whether providing the models with ground-truth error-type labels can improve refinement. Our results show that with this additional guidance, both models exhibit improved performance across all types, particularly in \emph{Mismatched}. These findings indicate that clarity regarding error types facilitates more effective LLM-based scene graph refinement.

\subsection{Do LLMs Hallucinate in Scene Graph Generation?}

Since hallucination is a known issue of LLMs, we investigate whether they hallucinate in the SA-SGG task. We consider hallucination occurred when any generated element is not present in the predefined set of elements, $L$. To account for instances in which the model simply copies elements that are explicitly mentioned in the provided description $d_i$, we further classify hallucinated outputs into two categories: (1) elements that are mentioned in the description but not included in $L$, and (2) entirely new elements not present in either $L$ or the description. As shown in Table~\ref{tab:hallucination}, Claude-3.5-Sonnet exhibits a very low incidence of hallucination, whereas smaller models tend to hallucinate more frequently.

\begin{table}[H]
\centering
\resizebox{0.9\columnwidth}{!}{%
\begin{tabular}{lccc}
\hline
\textbf{Model} & \textbf{Total} & \textbf{Desc.} & \textbf{New.} \\
\hline
Claude-3.5-Sonnet & 17 & 14 & 3 \\
GPT-4o & 215 & 156 & 59 \\
Qwen-2.5-7B & 395 & 192 & 203 \\
Mistral-7B & 616 & 235 & 381 \\
\hline
\end{tabular}
}
\caption{Hallucination counts in the SA-SGG task(1,188 data). "Total" indicates the total number of hallucinations, “Desc.” indicates “Description Elements,” which are explicit in the description but not in $L$, regarded as related but incorrect elements, and “New.” indicates “New Elements,” which are not included in either $L$ or the description, regarded as entirely unrelated elements.}
\label{tab:hallucination}
\end{table}

\section{Conclusion}
We present \ourbench{}, a benchmark for assessing large language models in scene graph understanding and generation. \ourbench{} comprises textual narratives and corresponding scene graphs to evaluate interpretation, reasoning, and structured representation derivation. Our experiments with 11 LLMs revealed moderate success in understanding tasks yet difficulties in generation tasks, with action decomposition posing the largest obstacle. We believe our comprehensive evaluation on LLMs’ ability to process scene graph representations paves the way for building more reliable systems in multimodal environments and enables future research on potential improvements.

\section*{Limitations} 
In the current study, both SGG tasks of \ourbench{} involve generating action-centric scene graphs from textual narratives.
However, the current textual narratives do not include object attributes (\eg, color, size), nor does \ourbench{} incorporate attributes in the scene graph generation process.
As \ourbench{} centers on a single actor, tasks do not involve additional individuals. Constructing a benchmark with more intricate scene graphs, including attributes and multiple actors, could provide more comprehensive insights. The primary objective of this study is to benchmark widely adopted LLMs for scene graph-related tasks, leveraging their knowledge and reasoning. Future directions may integrate multi-modal approaches, such as utilizing Vision-Language Models (VLMs) to derive textual narratives from images or videos.

\section*{Acknowledgements}
This work was supported by STEAM R\&D Project (RS-2024-00454458) and Global Young Connect Project (RS-2024-00407282), and Institute of Information \& Communications Technology Planning \& Evaluation (IITP) grant funded by the Korean government (MSIT)(No.RS-2020-II201361, Artificial Intelligence Graduate School Program (Yonsei University)). Jinyoung Yeo is a corresponding author.

\bibliography{anthology,custom}

\appendix
\section*{Appendix}

\section{Dataset Construction Details}
\begin{figure*}[htbp]
    \centering
    \includegraphics[width=0.9\linewidth]{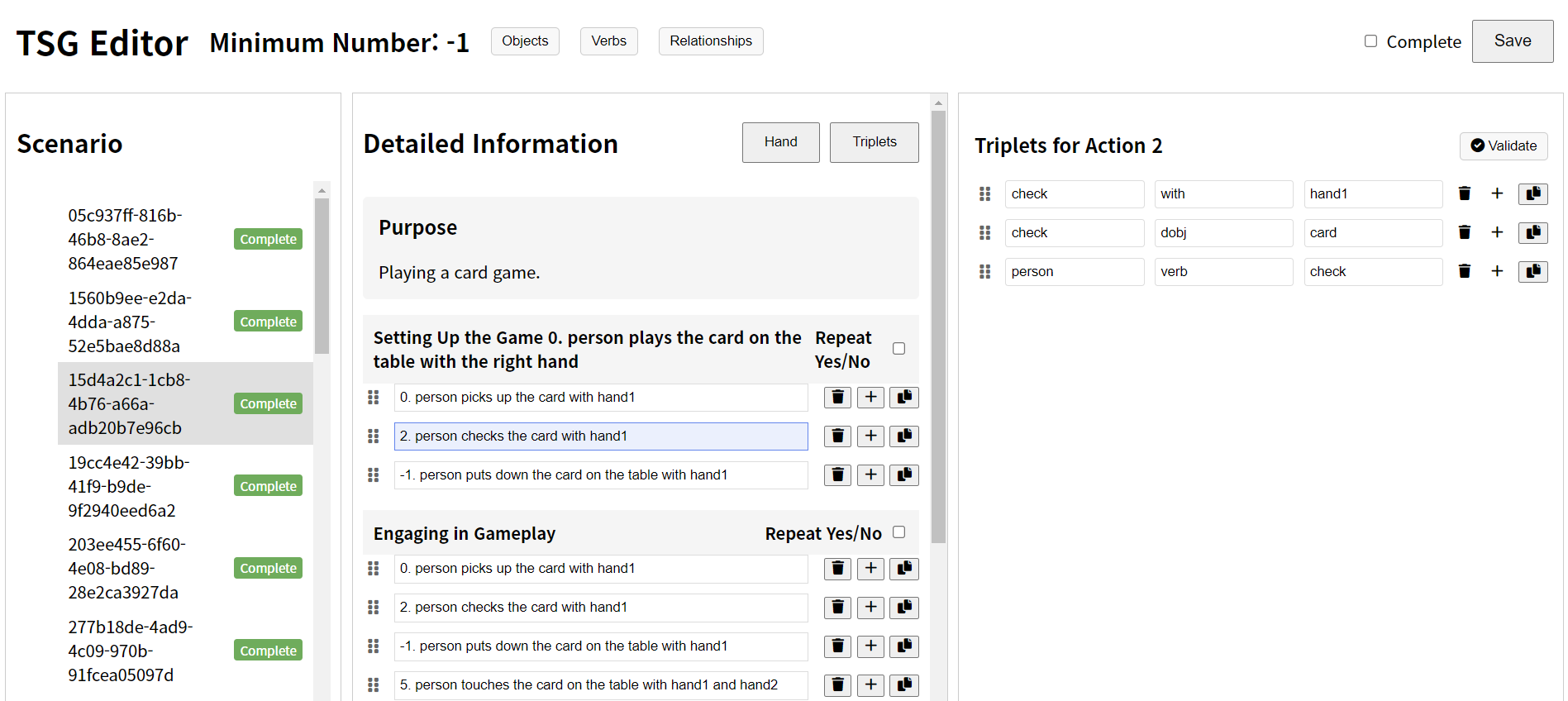}
    \caption{Example interface of the TSG Editor, designed to facilitate the creation of scene graph datasets from multiple scenarios. The left panel shows the list of scenarios with their completion statuses, while the right panel allows for specifying details such as action goals, hands used, objects, and relationship triplets.}
    \label{fig:tsg_editor}
\end{figure*}
\label{sec:appendix_data_construction}
We build upon the Ego-centric Action Scene Graphs (EASG) dataset~\citep{rodin2024action} to create a benchmark of our database and all four tasks. EASG provides spatio-temporal scene graphs derived from first-person video, where each graph encodes actions and objects from the perspective of the camera wearer. Although EASG dataset contains text annotations, they are simple verbalizations of the graph, without any contextual interaction between texts in a scenario. We instead aim to construct a database of human-like sentences, ensuring logical flow, contextuality, and complexity. For data construction, we perform three main steps, each serving a distinct purpose. All steps are done by a collaboration with an LLM and human annotators. Three annotators worked on human annotation, and they are paid on the hourly rate exceeding \$18, based on the estimated time required to complete the tasks. The screenshot of the annotation page is shown in Figure~\ref{fig:tsg_editor}.

\paragraph{Step 1: Narrative annotation.}
We utilize the scene graphs in EASG dataset, which cover diverse real world scenarios, to generate the initial natural language descriptions by prompting GPT-4o-mini. Then, human annotators remove redundant sentences, add missing necessary sentences, and fix unrealistic sentences (\eg, a single hand from holding multiple objects) considering the context.

\paragraph{Step 2: Scene graph annotation.}
Then, we generate scene graphs with GPT-4o based on each narrative resulted from Step 1. Here, we let the model generate the graph, guided by the predefined elements of the EASG dataset. The generated elements are then grouped by scenario (with each scenario corresponding to multiple scene graphs). Then, we let human workers review each instance, consisting of a text description, a scene graph, and scenario-specific elements. Following this review process, the scene graph and the scenario-specific elements—denoted as $L$—are finalized as the reference data.


\paragraph{Step 3: Text quality improvement.}
Although we have annotations for description and scene graph pairs, we additionally work on improving the quality of the text narratives, in terms of lexical diversity, coherency, and complexity. We prompt GPT-4o to paraphrase each sentence, given the preceding context, and also identify which word changes it has made. As this process may produce a description containing a similar element in the scenario-specific elements, which may yield a misleading annotation with the reference graph, we skip the instance. After paraphrasing, we prompt the model again to increase fluency regarding the context (\eg, adding a conjunction or modifier).

\paragraph{Annotator alignment process.}
To ensure consistency and quality across annotators, we underwent an alignment phase before each step. All annotators participated in a sampling-based alignment session under the supervision of a project lead. During this process, they were required to follow a set of detailed checklists specific to each step of the annotation workflow. These checklists served as shared standards to minimize variance and ensure high-quality outputs.


\vspace{3mm}
\noindent\textbf{Checklist for Step 1: Narrative Annotation}
\begin{itemize}
\item Sentences must accurately reflect the scene graph without exaggerating or downplaying its meaning.
\item Redundant sentences must be removed.
\item If the narrative lacks logical flow, new sentences should be added to resolve inconsistencies.
\item The order of usage for \texttt{hand1} and \texttt{hand2} must be strictly maintained.
\item Any newly added sentence must be expressible in the scene graph format.
\end{itemize}

\noindent\textbf{Checklist for Step 2: Scene Graph Annotation}
\begin{itemize}
\item All elements must be lemmatized.
\item All nodes must belong exactly to one of: \texttt{person}, \texttt{action}, \texttt{object}, or \texttt{hand}.
\item All edges must clearly be of type: \texttt{verb}, \texttt{dobj}, or \texttt{preposition}.
\item All major actions and relations described must be reflected in the scene graph without omission.
\item When a narrative includes compound actions, each action must be separated and represented by distinct scene graphs.
\item For repeated or sequential actions, actions must be represented in separate scene graphs reflecting order and frequency.
\item When describing grasping actions, the first object must be assigned to \texttt{hand1}, the second to \texttt{hand2}, and after all objects are released, the sequence restarts from \texttt{hand1}.
\item If synonyms are present in the predefined set of scenario-specific elements, only one representative term should be retained.
\item The set of scenario-specific elements may be expanded with additional elements in the graph, if they are deemed necessary based on the context.
\end{itemize}

\noindent\textbf{Checklist for Step 3: Text Quality Improvement}
\begin{itemize}
\item If the meaning of the original sentence is not preserved in paraphrased versions, the original sentence should be retained.
\item If the paraphrasing produces any different element that is present in the scenario-specific set, the original sentence should be retained, in order to avoid ambiguous annotation.
\item If the paraphrased sentence is not representable with the target scene graph format, the original sentence should be retained.
\item When multiple sentences are merged, the resulting sentence must comprehensively cover all the corresponding scene graphs. Unless, the original sentence should be retained.
\item Even if multiple sentences are implicitly expressed as a single sentence, it's acceptable as long as the original sentences can be logically inferred.
\end{itemize}

\paragraph{Data construction for understanding tasks.}
For generating questions for SGQA, we provide GPT-4o with the entire context in graph form to make multi-hop questions. We prompt it to make questions about common elements across multiple triplets (\eg, first object that had been grabbed), the state of nodes under specific conditions (\eg, before or after an action happened), etc.. For SGDS, we also prompt GPT-4o to perturb the candidates to generate sentences with similar words but that have different meanings with the reference. To finalize, human reviewers manually edit any incorrect question or answer to ensure accuracy.

\begin{figure}[ht]
    \centering
    \includegraphics[width=\columnwidth]{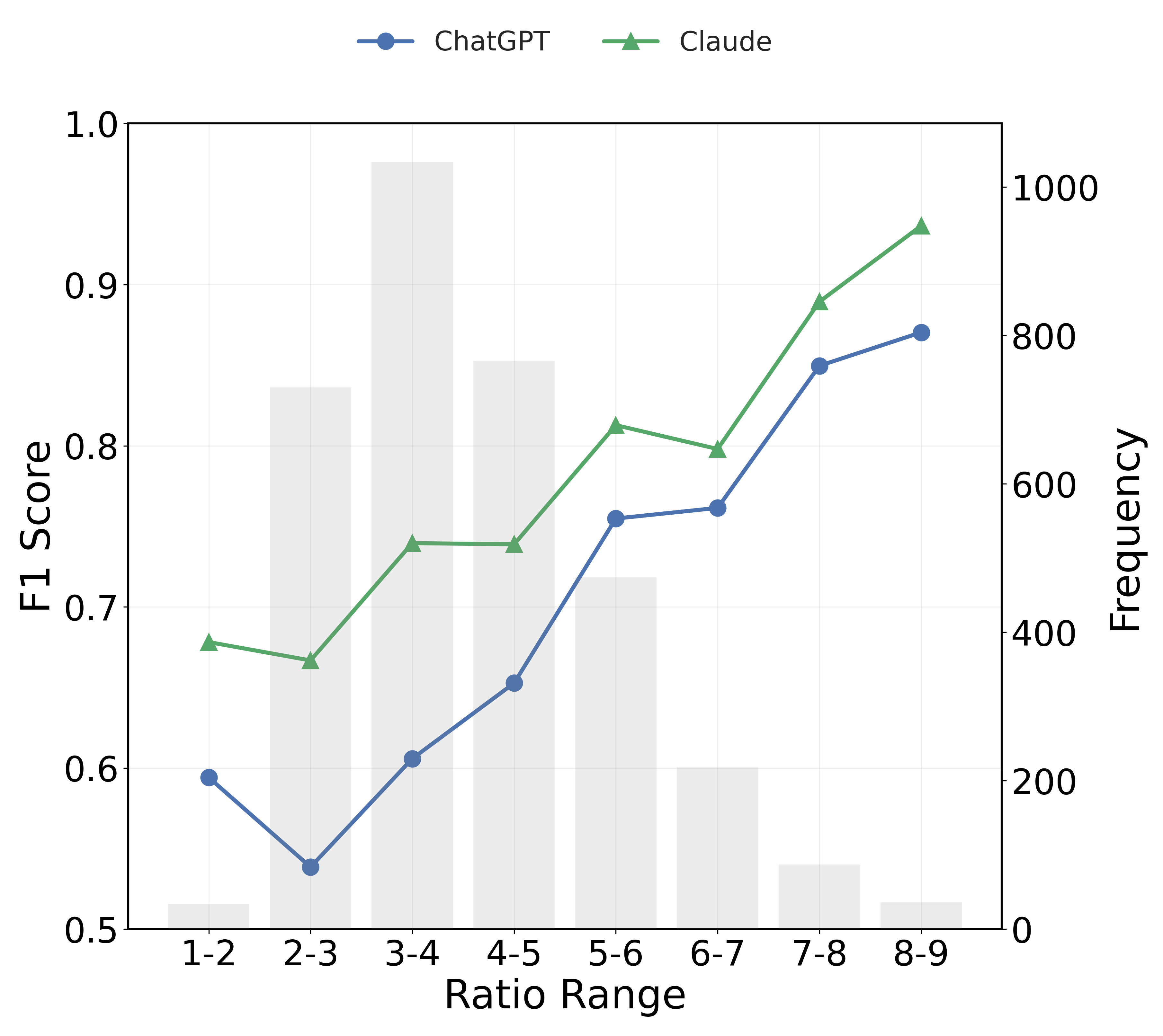}
    \caption{
The line graph shows $F_{1}$ scores of GPT-40 and Claude across varying levels of descriptiveness in the MA-SGG data. The histogram represents sample frequencies within each descriptiveness ratio range.
    }
    \label{fig:sentence-ratio}
\end{figure}

\section{Experiment Details}~\label{appendix:experiment_details}
We used OpenRouter (\url{https://openrouter.ai/}) for LLM prompting experiments. The temperature for models is set to 0.1 after a pilot study, and the scores in the experiment tables are results of running the inference only once. We also used NLTK(\url{https://www.nltk.org/}) for data processing, like lemmatization. The human performer is proficient in English and was instructed with task guidelines.

\section{Additional Analysis}
\subsection{Analysis of the Impact of Description Length on Scene Graph Generation}
We examined the effect of the descriptiveness of $d_i$ on scene graph generation performance. Using the MA-SGG data, we quantified descriptiveness as the ratio of the number of words in each description $d_i$ to the number of edges in the corresponding scene graphs $G_i$. A lower ratio indicates that the description is written concisely, whereas a higher ratio reflects a description with greater detail to express the scene. As shown in Figure~\ref{fig:sentence-ratio}, both GPT-4o and Claude-3.5-Sonnet perform better when provided with more detailed descriptions, which likely offer additional cues for constructing accurate scene graphs.

\subsection{Case Study for the MA-SGG Task}~\label{appendix:repetition_case}
As LLMs show low performance in the MA-SGG task, we show some failure cases in Table ~\ref{tab:failure_cases_appendix}.

\begin{table}[!h]
\centering
\small
\renewcommand{\arraystretch}{1}
\begin{tabular}{@{} p{0.95\linewidth} @{}}
\toprule 
\textbf{Case 1 }\\
\midrule
\textbf{[Context]} \\
 ... The wood was then positioned on the cardboard, ready for the next steps. The process began by picking up the stick. The stick was then put into the paint can and used to stir the paint thoroughly.\\
\noalign{\vskip 5pt}
\hdashline
\noalign{\vskip 5pt}
\textbf{[Description]} \\
This stirring step was repeated two more times to ensure the paint was well-mixed. \\
\midrule
\textbf{[Nodes in the Reference Graph]} \\
$V_1$: person, \textbf{stir}, stick, ...   \\
$V_2$: person, \textbf{stir}, stick, ...  \\

\midrule
\textbf{[Nodes in the Generated Graph]} \\
$V_1$: person, \textbf{stir}, stick, ...   \\
$V_2$: person, \textbf{paint}, stick, ...  \\

\bottomrule 
\toprule 
\textbf{Case 2} \\
\midrule 
\textbf{[Context]} \\
The painting process began by preparing the paintbrush, dipping it into the paint, and applying it to the railing.\\
\noalign{\vskip 5pt}
\hdashline
\noalign{\vskip 5pt}
\textbf{[Description]} \\
The railing was painted with the paintbrush, ensuring even coverage. This step was repeated three times to achieve a consistent finish. \\
\midrule
\textbf{[Nodes in the Reference Graph]} \\
$V_1$: person, \textbf{paint}, paintbrush, ...   \\
$V_2$: person, \textbf{paint}, paintbrush, ...  \\
$V_3$: person, \textbf{paint}, paintbrush, ...  \\

\midrule
\textbf{[Nodes in the Generated Graph]} \\
$V_1$: person, \textbf{pick-up}, paintbrush, ...   \\
$V_2$: person, \textbf{dip}, paintbrush, ...  \\
$V_3$: person, \textbf{paint}, paintbrush, ...  \\
\bottomrule

\end{tabular}
\caption{Failure cases in generating scene graphs for sequences of repetitive actions. \( V_j \) represents the set of nodes in the \( j \)-th graph corresponding to the description.}
\label{tab:failure_cases_appendix}
\end{table}

\subsection{Details for Performance}
This subsection provides supplemental information for two aspects. First, we evaluate the model’s ability to detect edges and nodes when explicit action is omitted (Edge and Node w/o Action Performance). Second, we present Action Segmentation Results using the Longest Common Subsequence (LCS) approach, reported in terms of Precision, Recall, and Action F1. The complete findings for these metrics can be found in Table ~\ref{table:edge_object_perf} and Table ~\ref{tab:action_results}.

\begin{table}[h]
    \centering
    \resizebox{0.48\textwidth}{!}{%
    \begin{tabular}{l c c c}
        \toprule
        Model & Precision & Recall & F1 Score \\
        \midrule
        Human              & 95.38 & 93.84 & 94.60 \\
        \midrule
        GPT-4o             & 67.45 & 85.24 & 73.15 \\
        GPT-4o-mini        & 56.43 & 73.35 & 61.22 \\
        Claude-3.5-Sonnet  & 74.58 & 85.12 & 77.85 \\
        Claude-3.5-Haiku   & 60.85 & 74.72 & 61.62 \\
        \midrule
        LLaMA-3.3-70B      & 63.33 & 76.45 & 66.37 \\
        Qwen-2.5-72B       & 68.02 & 75.45 & 68.67 \\
        DeepSeek-V3        & 57.11 & 78.28 & 62.93 \\
        Mixtral-8x22B      & 52.27 & 76.36 & 58.67 \\
        Mistral-large      & 60.37 & 73.76 & 63.48 \\
        \hdashline
        Qwen-2.5-7B        & 66.56 & 72.37 & 67.57 \\
        Mistral-7B         & 19.47 & 64.04 & 27.42 \\
        \bottomrule
    \end{tabular}
    }
    \caption{Action segmentation results using LCS for Precision, Recall, and Action F1.}
    \label{tab:action_results}
\end{table}

\begin{table*}[th]
    \centering
    \resizebox{\textwidth}{!}{
    \begin{tabular}{l ccc ccc ccc ccc}
        \toprule
        \multirow{2}{*}{Models} 
          & \multicolumn{6}{c}{Edge} 
          & \multicolumn{6}{c}{Node} \\
        \cmidrule(lr){2-7} \cmidrule(lr){8-13}
          & \multicolumn{3}{c}{Single} 
          & \multicolumn{3}{c}{Multiple} 
          & \multicolumn{3}{c}{Single} 
          & \multicolumn{3}{c}{Multiple} \\
        \cmidrule(lr){2-4} \cmidrule(lr){5-7} \cmidrule(lr){8-10} \cmidrule(lr){11-13}
          & Prec. & Rec. & F1 
          & Prec. & Rec. & F1 
          & Prec. & Rec. & F1
          & Prec. & Rec. & F1 \\
        \midrule
        GPT-4o
          & 86.88 & 82.90 & 84.42
          & 81.31 & 79.25 & 79.80
          & 83.15 & 80.76 & 81.70
          & 76.44 & 74.92 & 75.29 \\

        GPT-4o-mini
          & 53.44 & 52.72 & 52.88
          & 48.11 & 51.85 & 49.53
          & 78.53 & 77.28 & 77.53
          & 69.14 & 71.89 & 69.87 \\

        Claude-3.5-Sonnet
          & 93.46 & 93.92 & 93.64
          & 87.56 & 92.59 & 89.65
          & 87.65 & 88.71 & 87.87
          & 85.15 & 86.86 & 85.59 \\

        Claude-3.5-Haiku
          & 78.97 & 76.52 & 77.47
          & 73.04 & 75.06 & 73.72
          & 75.96 & 72.23 & 73.53
          & 67.78 & 70.54 & 68.61 \\

        \hdashline
        LLaMA-3.3-70B
          & 64.05 & 73.68 & 67.84
          & 63.62 & 73.28 & 67.42
          & 81.47 & 81.70 & 81.27
          & 75.65 & 80.61 & 77.11 \\

        Qwen-2.5-72B
          & 82.98 & 81.92 & 82.32
          & 84.53 & 82.19 & 79.74
          & 80.39 & 81.56 & 71.18
          & 74.74 & 74.75 & 70.68 \\

        DeepSeek-V3
          & 81.73 & 81.64 & 81.60
          & 85.95 & 77.26 & 82.28
          & 82.63 & 83.77 & 74.47
          & 77.77 & 77.77 & 72.93 \\

        Mixtral-8x22B
          & 62.26 & 64.22 & 62.87
          & 79.67 & 64.36 & 72.43
          & 68.11 & 75.13 & 54.43
          & 59.38 & 59.38 & 58.67 \\

        mistral-large-2411
          & 84.85 & 83.67 & 84.07
          & 83.70 & 83.90 & 84.57
          & 74.91 & 83.74 & 72.91
          & 73.16 & 73.16 & 69.87 \\
        \hdashline
        Qwen-2.5-7b
          & 19.59 & 18.17 & 18.58
          & 14.37 & 16.52 & 14.99
          & 77.80 & 72.74 & 74.29
          & 56.56 & 69.33 & 60.74 \\

        Mistral-7b
          & 37.21 & 35.97 & 36.09
          & 24.69 & 30.02 & 26.30
          & 52.94 & 50.85 & 50.10
          & 37.10 & 49.34 & 40.12 \\
        \bottomrule
    \end{tabular}
    }
    \caption{Edge and Node w/o Action Performance (\%)}
    \label{table:edge_object_perf}
\end{table*}

\subsection{Details for Data Distribution}
\begin{figure*}[htbp]
    \centering
    \includegraphics[width=0.9\linewidth]{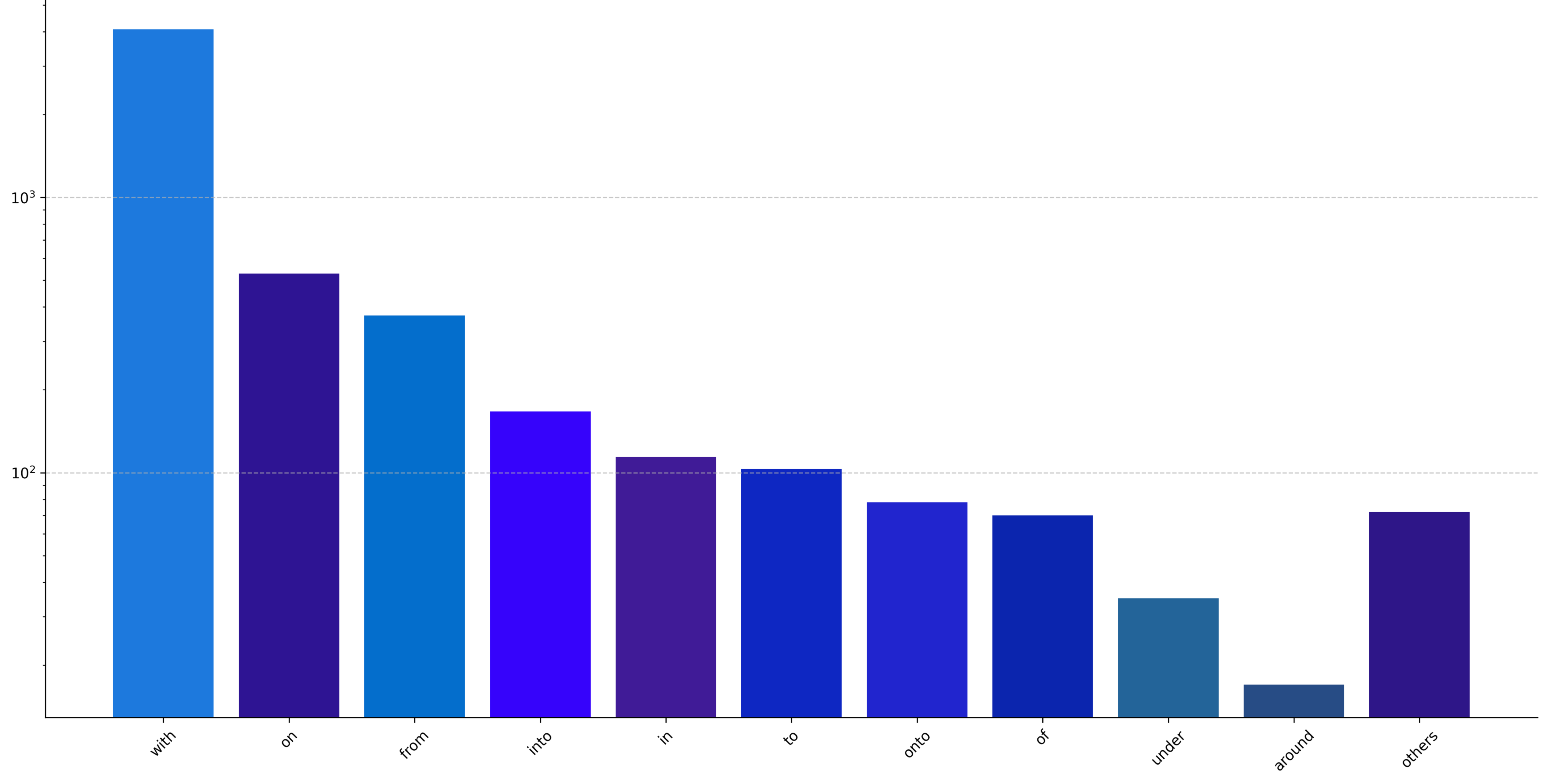}
    \caption{The distribution of relationships in the scene graph dataset, with “with” appearing most frequently, followed by “on” and “from.”}
    \label{fig:relationships_distribution}
\end{figure*}

\begin{figure*}[htbp]
    \centering
    \includegraphics[width=0.9\linewidth]{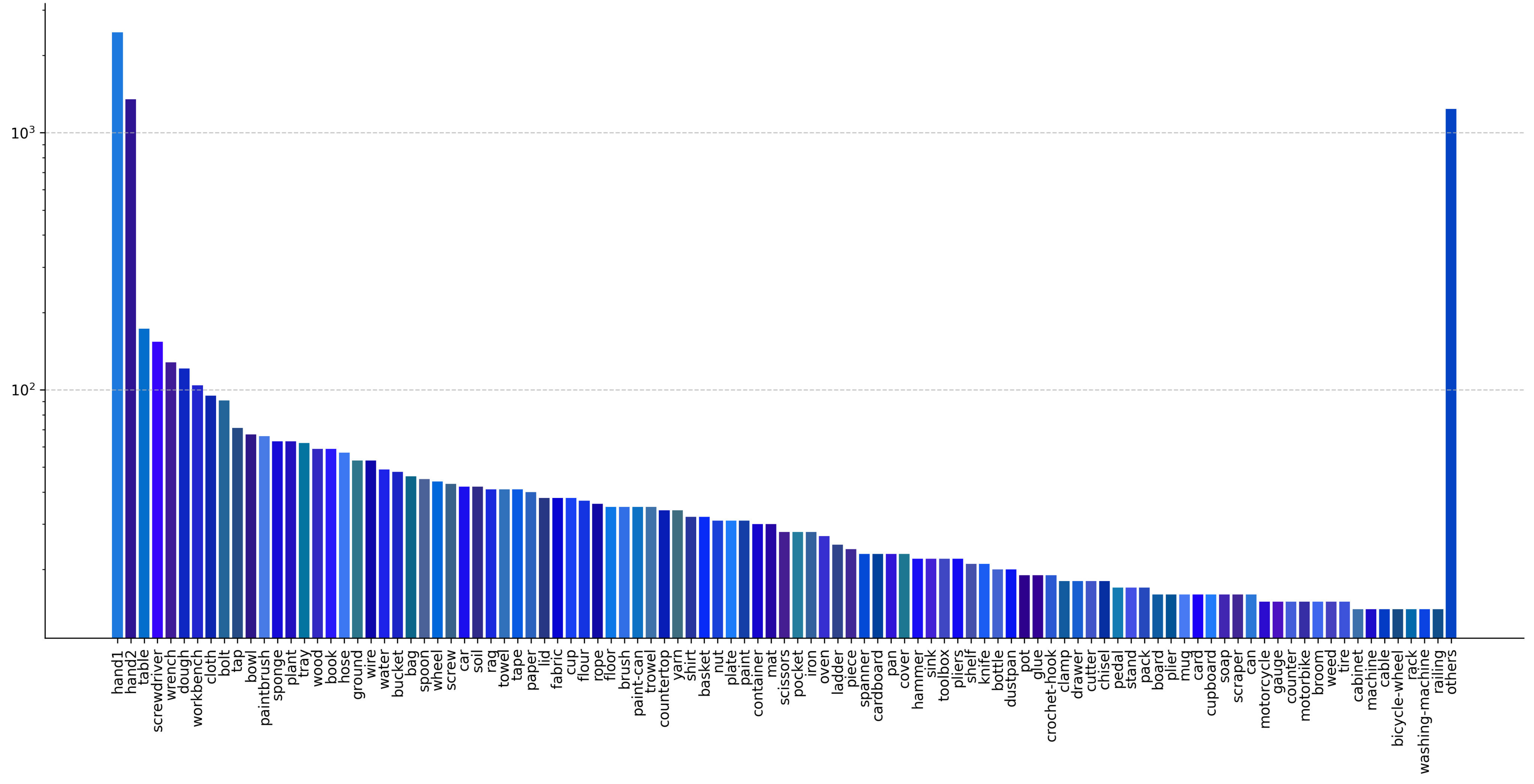}
    \caption{The frequency distribution of scene graph objects on a logarithmic scale, with “hand1” and “others” standing out as the most frequently occurring categories}
    \label{fig:objects_distribution}
\end{figure*}

\begin{figure*}[htbp]
    \centering
    \includegraphics[width=0.9\linewidth]{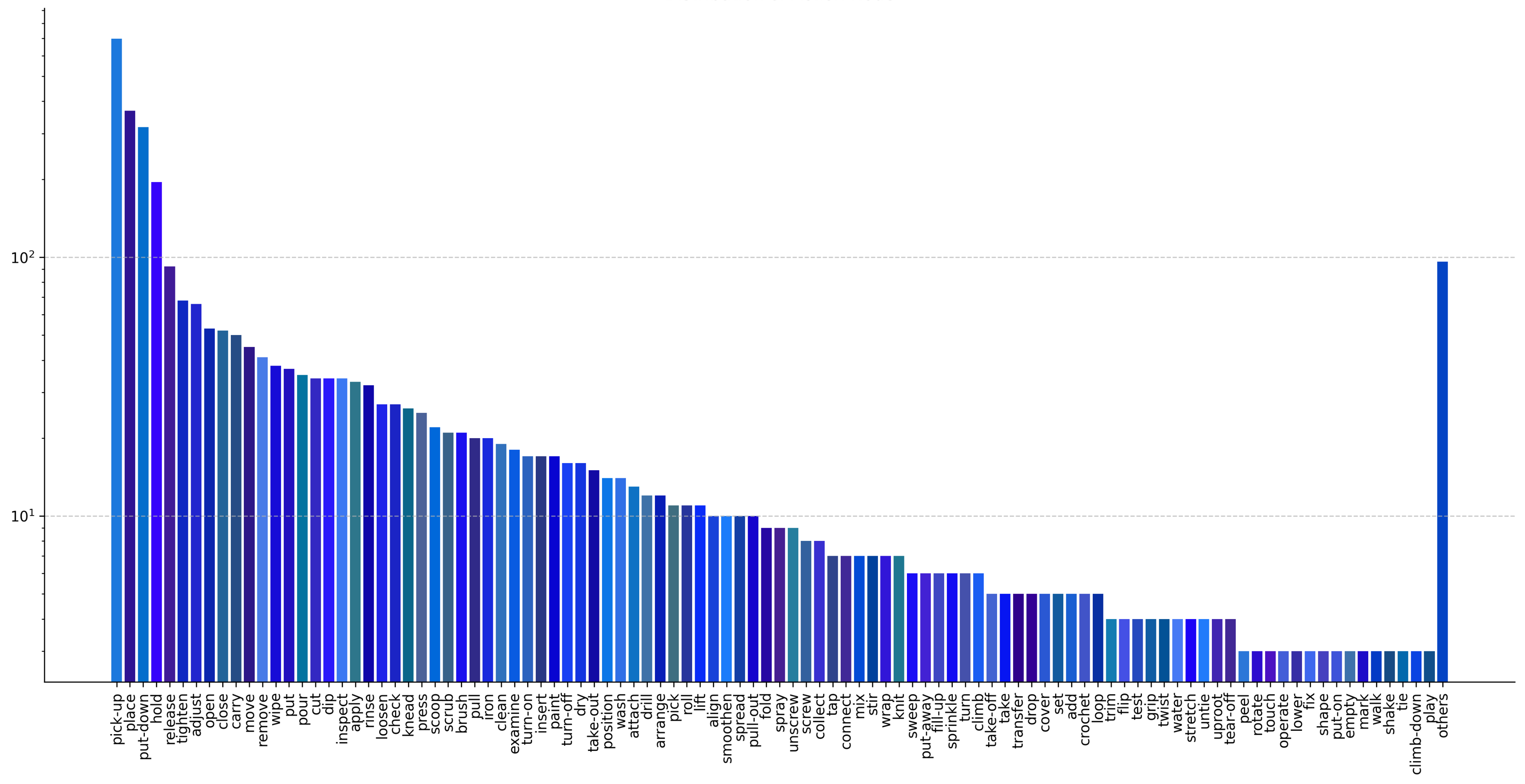}
    \caption{The frequency distribution of verbs in the scene graph, with “others” as the dominant category, followed by “pick up,” “place,” and “put down” among the most frequent specific actions.}
    \label{fig:verb_distribution}
\end{figure*}

The distribution of scene graph objects,  relationships and verbs in the dataset is visually summarized in Figure~\ref{fig:objects_distribution}, ~\ref{fig:relationships_distribution} and ~\ref{fig:verb_distribution}.

\section{Prompts}
\subsection{Zero-shot Prompts}
\label{appendix:eval_prompts}
This section provides full prompt. Each prompt is formulated to clearly specify input formats, rules for processing these inputs, and the exact manner in which the output should be generated.

\begin{itemize}
    \item \textbf{SGDS Task}: Figure~\ref{fig:base prompt for SGDS}
    \item \textbf{SGQA Task}: Figure~\ref{fig:base prompt for SGQA}
    \item \textbf{SA-SGG Task}: Figure~\ref{fig:base prompt for SA-SGG}
    \item \textbf{MA-SGG Task}: Figure~\ref{fig:base prompt for MA-SGG}
\end{itemize}

\subsection{Chain of Thought Prompts}
\label{appendix:CoT_prompts}
In addition to the base prompt, we incorporated a minimal form of Chain-of-Thought (CoT) prompting to guide the reasoning process.

\begin{itemize}
    \item \textbf{SGDS Task}: Figure~\ref{fig:CoT prompt for Scene Graph Description Selection}
    \item \textbf{SGQA Task}: Figure~\ref{fig:CoT prompt for Scene Graph Question Answering}
    \item \textbf{SA-SGG Task}: Figure~\ref{fig:CoT prompt for single graph generation}
    \item \textbf{MA-SGG Task}: Figure~\ref{fig:CoT prompt for multi graph generation}
\end{itemize}

\subsection{Few-shot Prompts}
\label{appendix:few_shot_prompts}
We employed a few-shot prompt with ten examples (10-shot), allowing the model to observe multiple instances of input-output pairs.
\begin{itemize}
    \item \textbf{SGDS Task}: Figure~\ref{fig:Fewshot prompt for Scene Graph DEscription Selection}
    \item \textbf{SGQA Task}: Figure~\ref{fig:Few-shot prompt for Scene Graph Qusetion Answering}
    \item \textbf{SA-SGG Task}: Figure~\ref{fig:Few-shot prompt for single graph generation}
    \item \textbf{MA-SGG Task}: Figure~\ref{fig:Few-shot prompt for multi graph generation}
\end{itemize}


\begin{figure*}[t!]
\small
\centering
\resizebox{0.86\textwidth}{!}{
\begin{tcolorbox}[
colframe=black,        
colback=gray!10,
arc=2mm,
boxrule=1.5pt,   
title=\textbf{Prompts},
fonttitle=\bfseries
]
    You are an AI that analyzes a Scene Graph based on the context and select the best text description of it among the given candidates.\\
    \\
    1. Input:\\
    \hspace{0.2cm}- Context: A list of scene graphs representing the preceding context.\\
    \hspace{0.2cm}- Each graph is composed of a set of triplets `[node1, edge, node2]`. `node1` and `node2` are one of person, action, object and hand. `edge` represents the relationship between them (e.g., `verb`, `dobj`, `from`, `with`).\\
    \hspace{0.2cm}- Target Scene Graph: A set of triplets that should be described into text correctly.\\
    \hspace{0.2cm}- Description Candidates: Candidates of sentence descriptions of the Target Scene Graph based on the Context.\\
    \\
    2. Task:\\
    \hspace{0.2cm}- Determine which description best matches the Target Scene Graph.\\
    \\
    3. Output:  \\
    \hspace{0.2cm}- Be sure to choose only one letter of the matching description.  \\
    \hspace{0.2cm}- Do not output any additional text or explanation. Only the letter in [ ] (e.g., [A]).\\
    \\
    Key rules of edges in a triplet:\\
    \hspace{0.2cm}- `verb` describes the action performed by `person`.\\
    \hspace{0.2cm}- `dobj` links the action to its direct object (`node2`).\\
    \hspace{0.2cm}- Other edges like `from` and `with` describe spatial relationships between nodes.\\
    \\
    Input:\\
    - Context: \{context\}\\
    - Target Scene Graph: \{triplet\}\\
    - Description Candidates: \\
    \{sentences\}
    
\end{tcolorbox}
}
\caption{The zero-shot prompt for scene graph description selection tasks.}
\label{fig:base prompt for SGDS}
\end{figure*}
\begin{figure*}[t!]
\small
\centering
\resizebox{0.86\textwidth}{!}{
\begin{tcolorbox}[
colframe=black,        
colback=gray!10,
arc=2mm,
boxrule=1.5pt,   
title=\textbf{Prompts},
fonttitle=\bfseries
]
    You are a highly advanced language model specialized in answering questions based on a given scene graph and question. Your task is to analyze the scene graph and provide the correct answer in a single word. Your output must strictly follow the format [answer], and nothing else should be printed. Ensure that your answer is concise, accurate, and matches the format exactly.
    \\
    \\
    Scene Graph: \{scene\_graph\} \\
    Question: \{question\} \\
\end{tcolorbox}
}
\caption{The zero-shot prompt for scene graph question answering tasks.}
\label{fig:base prompt for SGQA}
\end{figure*}
\begin{figure*}[t!]
\small
\centering
\resizebox{0.86\textwidth}{!}{
\begin{tcolorbox}[
colframe=black,        
colback=gray!10,
arc=2mm,
boxrule=1.5pt,   
title=\textbf{Prompts},
fonttitle=\bfseries
]
    You are an AI model tasked with generating a scene graph based on a given sentence, adhering to specific rules for the graph, nodes, and edges, while considering the provided context, available nodes, and available edges.
    \\ \\
    Rules for Scene Graph Representation:
    \\1. A graph is composed of one or more triplets of nodes and edges.
    \\2. A triplet starts with a node and another node is connected by an edge. (Format: node -> edge -> node)
    \\3. Each triplet is split with a new line.
    \\4. There must be a triplet that starts with a person node.
    \\5. All nodes and edges must be one of "Available nodes" or "Available edges" provided.
    \\ \\
    Rules for Node:
    \\1. A node can be person, any action, any object, or any hand.
    \\2. A node may appear explicitly or be hidden implicitly in the given sentence. Consider the context to identify the node.
    \\3. Map synonyms or semantically similar words to nodes in the "Available nodes" list.
    \\4. Use default tools or body parts for actions that imply them (e.g., hands for grasping).
    \\5. Include "person" as the starting node in the graph.
    \\ \\
    Special Rules for Hand Node:
    \\1. If both hands are empty and a node is grasped, represent it as "hand1."
    \\2. If one hand holds a node and another node is grasped, represent it as "hand2."
    \\3. If all hands release their objects, reset the next grasping hand to "hand1."
    \\4. Ensure "hand1" and "hand2" are used contextually to avoid overlap or ambiguity.
    \\5. If the sentence implies using both hands (e.g., lifting a large object), represent both hands explicitly (e.g., hand1, hand2).
    \\ \\
    Rules for Edge:
    \\1. An edge can be verb, dobj, or any preposition.
    \\2. Map synonyms or semantically similar words to edges in the "Available edges" list.
    \\1. verb: can only connect person and an action node. (e.g., person -> verb -> add)
    \\2. dobj: connects an action and an object node, only when it is the direct object of the action (e.g., add -> dobj -> flour)
    \\3. preposition: connects one of the four types of node pairs: action \& object / action \& hand / object \& object / hand \& object (e.g., take -> from -> table)
    \\ \\
    Output Format:
    The output must consist of triplets (one per line) in the format below.
    node -> edge -> node
    node -> edge -> node
    ...
    \\ \\
    Use only the "Available nodes" and "Available edges" provided. No additional text, explanations, or formatting should be included.\\
    
    Inputs:\\ 
    Context: \{context\}\\
    Target sentence: \{target\_sentence\}\\
    Available nodes: \{available\_nodes\}\\
    Available edges: \{available\_edges\}
\end{tcolorbox}
}
\caption{The zero-shot prompt for single action scene graph generation tasks.}
\label{fig:base prompt for SA-SGG}
\end{figure*}
\begin{figure*}[t!]
\small
\centering
\resizebox{0.86\textwidth}{!}{
\begin{tcolorbox}[
colframe=black,        
colback=gray!10,
arc=2mm,
boxrule=1.5pt,   
title=\textbf{Prompts},
fonttitle=\bfseries
]
    You are an AI model tasked with generating scene graphs based on a given sentence. Your goal is to create exactly the specified number of scene graphs by extracting meaningful relationships between entities, actions, and objects while ensuring that the scene graphs represent actions that would visually appear in a scene.
    \\ \\
    Rules for Generating Multiple Scene Graphs:\\
    1. Generate precisely \{num\_scene\_graphs\} scene graphs—no more, no less.\\
    2. Each scene graph must depict an action that would be explicitly visible in a scene.\\
    3. If the sentence contains multiple implicit actions, distribute them among the scene graphs while ensuring the total count matches \{num\_scene\_graphs\}.\\
    4. If there are fewer visible actions than
    \{num\_scene\_graphs\}, **additional relevant actions may be inferred** to reach the required count.\\
    5. However, use only the "Available Nodes" and "Available Edges" provided. **If a necessary node is missing, use the closest semantically matching node from the available list**.\\
    6. Ensure each graph maintains logical coherence while including essential contextual elements.\\
    \\
    Rules for A Scene Graph Representation:\\
    1. A graph is composed of one or more triplets of nodes and edges.\\
    2. A triplet starts with a node and another node is connected by an edge. (Format: node -> edge -> node)\\
    3. Each triplet is split with a new line.\\
    4. There must be exactly one triplet that starts with a person node in a graph.\\
    5. All nodes and edges must be one of "Available nodes" or "Available edges" provided.\\
    \\
    Rules for Node:\\
    1. A node can be person, any action, any object, or any hand.\\
    2. A node may appear explicitly or be hidden implicitly in the given sentence. Consider the context to identify the node from the "Available nodes" list, but do not create a new one.\\
    3. Map synonyms or semantically similar words to nodes in the "Available nodes" list.\\
    4. Use default tools or body parts for actions that imply them (e.g., hands for grasping).\\
    5. Treat each action as a node.\\
    6. Include "person" as the starting node in the graph.
    \\ \\
    Special Rules for Hand Node:\\
    1. If both hands are empty and a node is grasped, represent it as "hand1."\\
    2. If one hand holds a node and another node is grasped, represent it as "hand2."\\
    3. If all hands release their objects, reset the next grasping hand to "hand1."\\
    4. Ensure "hand1" and "hand2" are used contextually to avoid overlap or ambiguity.\\
    5. If the sentence implies using both hands (e.g., lifting a large object), represent both hands explicitly (e.g., hand1, hand2).\\
    \\
    Rules for Edge:\\
    1. An edge can be verb, dobj, or any preposition.\\
    2. Use only the edges listed under "Available edges."\\
    3. Here are the explanations for each edge.\\
      - verb: can only connect person and an action node. (e.g., person -> verb -> add)
      - dobj: connects an action and an object node, only when it is the direct object of the action (e.g., add -> dobj -> flour)
      - preposition: connects one of the four types of node pairs: action \& object / action \& hand / object \& object / hand \& object (e.g., take -> from -> table)
    \\ \\
    Output Format:\\
    The output must consist of exactly \{num\_scene\_graphs\} scene graphs, each separated with a blank line. For a graph, output one triplet per line. Follow the format below (an example of three scene graphs of multiple triplets):\\
    node -> edge -> node\\
    node -> edge -> node\\
    ...\\ \\
    node -> edge -> node\\
    node -> edge -> node\\
    ...\\ \\
    node -> edge -> node\\
    node -> edge -> node\\
    ...\\ \\
    \\Use only the "Available Nodes" and "Available Edges" provided. No additional text, explanations, or formatting should be included.\\
    
    Inputs:\\
    - Context: \{context\}\\
    - Target sentence: \{target\_sentence\}\\
    - Available nodes: \{available\_nodes\}\\
    - Available edges: \{available\_edges\}\\
    - Number of Scene Graphs: \{num\_scene\_graphs\}
\end{tcolorbox}
}
\caption{The zero-shot prompt for multiple action scene graph generation tasks.}
\label{fig:base prompt for MA-SGG}
\end{figure*}

\begin{figure*}[t!]
\small
\centering
\resizebox{0.86\textwidth}{!}{
\begin{tcolorbox}[
colframe=black,        
colback=gray!10,
arc=2mm,
boxrule=1.5pt,   
title=\textbf{Prompts},
fonttitle=\bfseries
]
    You are an AI that analyzes a Scene Graph based on the context and select the best text description of it among the given candidates.\\

    1. Input:\\
    \hspace{0.2cm}- Context: A list of scene graphs representing the preceding context.\\
    \hspace{0.2cm}\hspace{0.2cm}- Each graph is composed of a set of triplets [node1, edge, node2]. `node1` and `node2` are one of person, action, object and hand. `edge` represents the relationship between them (e.g., `verb`, `dobj`, `from`, `with`).\\
    \hspace{0.2cm}- Target Scene Graph: A set of triplets that should be described into text correctly.\\
    \hspace{0.2cm}- Description Candidates: Candidates of sentence descriptions of the Target Scene Graph based on the Context.\\
    
    2. Task:\\
    \hspace{0.2cm}- Think step-by-step and determine which description best matches the Target Scene Graph.\\
    
    3. Output:\\
    \hspace{0.2cm}- Output your rationale under "Think:"\\
    \hspace{0.2cm}- Then, output your final answer under "Final Answer:"\\
    \hspace{0.2cm}- For the final answer, be sure to choose only one letter of the matching description and write it in the format of (e.g., [X]), where "X" represents a single alphabet letter.\\
    
    Key rules of edges in a triplet:\\
    \hspace{0.2cm}- `verb` describes the action performed by `person`.\\
    \hspace{0.2cm}- `dobj` links the action to its direct object (`node2`).\\
    \hspace{0.2cm}- Other edges like `from` and `with` describe spatial relationships between nodes.\\
    
    Input:\\
    - Context: \{context\}\\
    - Target Scene Graph: \{triplet\}\\
    - Description Candidates:\\
    \{sentences\}\\
    
    Now, think step-by-step and output the final answer.\\
    Think:
    
\end{tcolorbox}
}
\caption{CoT prompt for scene graph description selection tasks.}
\label{fig:CoT prompt for Scene Graph Description Selection}
\end{figure*}
\begin{figure*}[t!]
\small
\centering
\resizebox{0.86\textwidth}{!}{
\begin{tcolorbox}[
colframe=black,        
colback=gray!10,
arc=2mm,
boxrule=1.5pt,   
title=\textbf{Prompts},
fonttitle=\bfseries
]
    You are a highly advanced language model specialized in answering questions based on a given scene graph and a question. Your task is to analyze the scene graphs and provide the correct answer in a single word. Think step-by-step under "Think:", and generate the final answer under "Final Answer:". Ensure that your final answer is a single word in the triplet, and do not generate additional explanations after it.\\

    Scene Graph: \{scene\_graph\}\\
    Question: \{question\}\\
    
    Think:
\end{tcolorbox}
}
\caption{CoT prompt for scene graph question answering tasks.}
\label{fig:CoT prompt for Scene Graph Question Answering}
\end{figure*}
\begin{figure*}[t!]
\small
\centering
\resizebox{0.86\textwidth}{!}{
\begin{tcolorbox}[
colframe=black,        
colback=gray!10,
arc=2mm,
boxrule=1.5pt,   
title=\textbf{Prompts},
fonttitle=\bfseries
]
    You are an AI model tasked with generating a scene graph based on a given sentence, adhering to specific rules for the graph, nodes, and edges, while considering the provided context, available nodes, and available edges. Think step-by-step and generate the graph in triplets. (Stick only to the target sentence and avoid over-predicting the next scene.)\\

Rules for Scene Graph Representation:\\
1. A graph is composed of one or more triplets of nodes and edges.\\
2. A triplet starts with a node and another node is connected by an edge. (Format: node -\textgreater{} edge -\textgreater{} node)\\
3. Each triplet is split with a new line.\\
4. There must be a triplet that starts with a person node.\\
5. All nodes and edges must be one of "Available nodes" or "Available edges" provided.\\

Rules for Node:\\
1. A node can be person, any action, any object, or any hand.\\
2. A node may appear explicitly or be hidden implicitly in the given sentence. Consider the context to identify the node.\\
3. Map synonyms or semantically similar words to nodes in the "Available nodes" list.\\
4. Use default tools or body parts for actions that imply them (e.g., hands for grasping).\\
5. Include "person" as the starting node in the graph.\\

Special Rules for Hand Node:\\
1. If both hands are empty and a node is grasped, represent it as "hand1."\\
2. If one hand holds a node and another node is grasped, represent it as "hand2."\\
3. If all hands release their objects, reset the next grasping hand to "hand1."\\
4. Ensure "hand1" and "hand2" are used contextually to avoid overlap or ambiguity.\\
5. If the sentence implies using both hands (e.g., lifting a large object), represent both hands explicitly (e.g., hand1, hand2).\\

Rules for Edge:\\
1. An edge can be verb, dobj, or any preposition.\\
2. Use only the edges listed under "Available edges."\\
3. Here are the explanations for each edge.\\
  - verb: can only connect person and an action node. (e.g., person -\textgreater{} verb -\textgreater{} add)\\
  - dobj: connects an action and an object node, only when it is the direct object of the action (e.g., add -\textgreater{} dobj -\textgreater{} flour)\\
  - preposition: connects one of the four types of node pairs: action \& object / action \& hand / object \& object / hand \& object (e.g., take -\textgreater{} from -\textgreater{} table)\\

    Output Format:\\
    The output must consist of your rationale and triplets (one per line) in the format below.\\
    Think:\\
    (Write your rationale here, but do not predict the next scene)\\
    
    Scene Graph:\\
    node -\textgreater{} edge -\textgreater{} node\\
    node -\textgreater{} edge -\textgreater{} node\\
    ...\\
    
    Use only the "Available nodes" and "Available edges" provided, and follow the format correctly. After generating the scene graph, no additional text or explanations should be included.\\
    
    Inputs:\\
    Context: \{context\}\\
    Target sentence: \{target\_sentence\}\\
    Available nodes: \{available\_nodes\}\\
    Available edges: \{available\_edges\}\\
    
    Think:
    
\end{tcolorbox}
}
\caption{CoT prompt for single action scene graph generation tasks.}
\label{fig:CoT prompt for single graph generation}
\end{figure*}
\begin{figure*}[t!]
\small
\centering
\resizebox{0.86\textwidth}{!}{
\begin{tcolorbox}[
colframe=black,        
colback=gray!10,
arc=2mm,
boxrule=1.5pt,   
title=\textbf{Prompts},
fonttitle=\bfseries
]
    You are an AI model tasked with generating scene graphs based on a given sentence. Your goal is to create exactly the specified number of scene graphs by extracting meaningful relationships between entities, actions, and objects while ensuring that the scene graphs represent actions that would visually appear in a scene. Read the rules below and think step-by-step to generate correct scene graphs that represent the target sentence.\\

    Rules for Generating Multiple Scene Graphs:\\
    1. Generate precisely \{num\_scene\_graphs\} scene graphs---no more, no less.\\
    2. Each scene graph must depict an action that would be explicitly visible in a scene.\\
    3. If the sentence contains multiple implicit actions, distribute them among the scene graphs while ensuring the total count matches \{num\_scene\_graphs\}.\\
    4. If there are fewer visible actions than \{num\_scene\_graphs\}, \*\*additional relevant actions may be inferred\*\* to reach the required count.\\
    5. However, use only the "Available Nodes" and "Available Edges" provided. \*\*If a necessary node is missing, use the closest semantically matching node from the available list\*\*.\\
    6. Ensure each graph maintains logical coherence while including essential contextual elements.\\
    
    Rules for A Scene Graph Representation:\\
    1. A graph is composed of one or more triplets of nodes and edges.\\
    2. A triplet starts with a node and another node is connected by an edge. (Format: node -\textgreater{} edge -\textgreater{} node)\\
    3. Each triplet is split with a new line.\\
    4. There must be exactly one triplet that starts with a person node in a graph.\\
    5. All nodes and edges must be one of "Available nodes" or "Available edges" provided.\\
    
    Rules for Node:\\
    1. A node can be person, any action, any object, or any hand.\\
    2. A node may appear explicitly or be hidden implicitly in the given sentence. Consider the context to identify the node from the "Available nodes" list, but do not create a new one.\\
    3. Map synonyms or semantically similar words to nodes in the "Available nodes" list.\\
    4. Use default tools or body parts for actions that imply them (e.g., hands for grasping).\\
    5. Treat each action as a node.\\
    6. Include "person" as the starting node in the graph.\\
    
    Special Rules for Hand Node:\\
    1. If both hands are empty and a node is grasped, represent it as "hand1."\\
    2. If one hand holds a node and another node is grasped, represent it as "hand2."\\
    3. If all hands release their objects, reset the next grasping hand to "hand1."\\
    4. Ensure "hand1" and "hand2" are used contextually to avoid overlap or ambiguity.\\
    5. If the sentence implies using both hands (e.g., lifting a large object), represent both hands explicitly (e.g., hand1, hand2).\\
    
    Rules for Edge:\\
    1. An edge can be verb, dobj, or any preposition.\\
    2. Use only the edges listed under "Available edges."\\
    3. Here are the explanations for each edge.\\
      - verb: can only connect person and an action node. (e.g., person -\textgreater{} verb -\textgreater{} add)\\
      - dobj: connects an action and an object node, only when it is the direct object of the action (e.g., add -\textgreater{} dobj -\textgreater{} flour)\\
      - preposition: connects one of the four types of node pairs: action \& object / action \& hand / object \& object / hand \& object (e.g., take -\textgreater{} from -\textgreater{} table)\\
    
    Output Format:\\
    The output must consist of your rationale and exactly \{num\_scene\_graphs\} scene graphs, each separated with a blank line. For a graph, output one triplet per line. Follow the format below (an example of three scene graphs of multiple triplets):\\
    Think:\\
    (Write your rationale here)\\
    
    node -\textgreater{} edge -\textgreater{} node\\
    node -\textgreater{} edge -\textgreater{} node\\
    ...\\
    
    node -\textgreater{} edge -\textgreater{} node\\
    node -\textgreater{} edge -\textgreater{} node\\
    ...\\
    
    node -\textgreater{} edge -\textgreater{} node\\
    node -\textgreater{} edge -\textgreater{} node\\
    ...\\
    
    Use only the "Available Nodes" and "Available Edges" provided. After generating the scene graphs, no additional text or explanations should be included.\\
    
    Inputs:\\
    - Context: \{context\}\\
    - Target sentence: \{target\_sentence\}\\
    - Available nodes: \{available\_nodes\}\\
    - Available edges: \{available\_edges\}\\
    - Number of Scene Graphs: \{num\_scene\_graphs\}\\
    Think:

\end{tcolorbox}
}
\caption{CoT prompt for multiple action scene graph generation tasks.}
\label{fig:CoT prompt for multi graph generation}
\end{figure*}

\begin{figure*}[t!]
\small
\centering
\resizebox{0.86\textwidth}{!}{
\begin{tcolorbox}[
colframe=black,        
colback=gray!10,
arc=2mm,
boxrule=1.5pt,   
title=\textbf{Prompts},
fonttitle=\bfseries
]
    You are an AI that analyzes a Scene Graph based on the context and select the best text description of it among the given candidates.\\
    
    1. Input:\\
    \hspace{0.2cm}- Context: A list of scene graphs representing the preceding context.\\
    \hspace{0.2cm}\hspace{0.2cm}- Each graph is composed of a set of triplets [node1, edge, node2]. `node1` and `node2` are one of person, action, object and hand. `edge` represents the relationship between them (e.g., `verb`, `dobj`, `from`, `with`).\\
    \hspace{0.2cm}- Target Scene Graph: A set of triplets that should be described into text correctly.\\
    \hspace{0.2cm}- Description Candidates: Candidates of sentence descriptions of the Target Scene Graph based on the Context.\\
    
    2. Task:\\
    \hspace{0.2cm}- Determine which description best matches the Target Scene Graph.\\
    
    3. Output:\\
    \hspace{0.2cm}- Be sure to choose only one letter of the matching description.\\
    \hspace{0.2cm}- Do not output any additional text or explanation. Only the letter in [ ] (e.g., [A]).\\
    
    Key rules of edges in a triplet:\\
    \hspace{0.2cm}- `verb` describes the action performed by `person`.\\
    \hspace{0.2cm}- `dobj` links the action to its direct object (`node2`).\\
    \hspace{0.2cm}- Other edges like `from` and `with` describe spatial relationships between nodes.\\
    
    Example Input 1:\\
    \hspace{0.2cm}- Context: [[['arrange', 'with', 'hand1'], ['arrange', 'with', 'hand2'], ['arrange', 'dobj', 'book'], ['person', 'verb', 'arrange']], [['put', 'dobj', 'book'], ['put', 'on', 'floor'], ['put', 'with', 'hand1'], ['put', 'with', 'hand2'], ['person', 'verb', 'put']], [['put', 'dobj', 'book'], ['put', 'on', 'bookshelf'], ['put', 'with', 'hand1'], ['put', 'with', 'hand2'], ['person', 'verb', 'put']]]\\
    \hspace{0.2cm}- Target Scene Graph: [['align', 'with', 'hand1'], ['align', 'with', 'hand2'], ['align', 'dobj', 'book'], ['align', 'on', 'bookshelf'], ['person', 'verb', 'align']]\\
    \hspace{0.2cm}- Description Candidates:\\
    \hspace{0.2cm}A: Finally, the flask was set down.\\
    \hspace{0.2cm}B: It was then aligned neatly on the shelf using both hands.\\
    \hspace{0.2cm}C: After achieving the desired consistency, the stick was removed from the paint can.\\
    \hspace{0.2cm}D: The cord was then sliced using the cord cutter.\\
    \hspace{0.2cm}E: Once the massaging was complete, the batter was positioned back on the counter with both hands.\\
    
    Example Output 1:\\
    \hspace{0.2cm}[B]\\
    
    ...\\
    
    Example Input 10:\\
    \hspace{0.2cm}- Context: [[['pick-up', 'with', 'hand1'], ['pick-up', 'with', 'hand2'], ['pick-up', 'dobj', 'rope'], ['person', 'verb', 'pick-up']], [['tie', 'with', 'hand1'], ['tie', 'with', 'hand2'], ['tie', 'dobj', 'rope'], ['tie', 'around', 'plant'], ['person', 'verb', 'tie']]]\\
    \hspace{0.2cm}- Target Scene Graph: [['pull', 'with', 'hand1'], ['pull', 'with', 'hand2'], ['pull', 'dobj', 'rope'], ['person', 'verb', 'pull'], ['pull', 'to', 'tighten']]\\
    \hspace{0.2cm}- Description Candidates:\\
    \hspace{0.2cm}A: The rope was then pushed loose with both hands to ensure a relaxed hold.\\
    \hspace{0.2cm}B: The rope was then pulled slack with one hand to ensure a loose grip.\\
    \hspace{0.2cm}C: The rope was then pulled tight with both hands to ensure a firm grip.\\
    \hspace{0.2cm}D: The rope was then dropped with both hands to ensure it stayed untightened.\\
    \hspace{0.2cm}E: The rope was then pulled apart with no hands to ensure it remained loose.\\
    
    Example Output 10:\\
    \hspace{0.2cm}[C]\\
    
    Input:\\
    \hspace{0.2cm}- Context: \{context\}\\
    \hspace{0.2cm}- Target Scene Graph: \{triplet\}\\
    \hspace{0.2cm}- Description Candidates:\\
    \{sentences\}\\
    
    Output:
    
\end{tcolorbox}
}
\caption{Few-shot prompt for scene graph description selection tasks.}
\label{fig:Fewshot prompt for Scene Graph DEscription Selection}
\end{figure*}
\begin{figure*}[t!]
\small
\centering
\resizebox{0.86\textwidth}{!}{
\begin{tcolorbox}[
colframe=black,        
colback=gray!10,
arc=2mm,
boxrule=1.5pt,   
title=\textbf{Prompts},
fonttitle=\bfseries
]
    You are a highly advanced language model specialized in answering questions based on a given scene graph and question. Your task is to analyze the scene graph and provide the correct answer in a single word. Your output must strictly follow the format [answer], and nothing else should be printed. Ensure that your answer is concise, accurate, and matches the format exactly.\\
    
    Example Input 1:\\
    \hspace{0.2cm}Scene Graph: [[["person", "verb", "pick-up"], ["pick-up", "dobj", "screw"], ["pick-up", "from", "bowl"], ["pick-up", "with", "hand2"]], [["person", "verb", "position"], ["position", "dobj", "screw"], ["position", "on", "furniture-piece"], ["position", "with", "hand2"]]]\\
    \hspace{0.2cm}Question: What object was positioned immediately after being picked up from the bowl?\\
    
    Example Output 1:\\
    \hspace{0.2cm}[screw]\\
    
    ...\\
    
    Example Input 10:\\
    \hspace{0.2cm}Scene Graph: [[["person", "verb", "pick-up"], ["pick-up", "with", "hand1"], ["pick-up", "dobj", "mop-stick"]], [["person", "verb", "sweep"], ["sweep", "with", "hand1"], ["sweep", "with", "hand2"], ["sweep", "with", "mop-stick"], ["sweep", "dobj", "floor"], ["sweep", "in", "car"]], [["person", "verb", "close"], ["close", "with", "hand1"], ["close", "dobj", "door"]], [["person", "verb", "place"], ["place", "with", "hand2"], ["place", "dobj", "mop-stick"], ["place", "on", "floor"]], [["person", "verb", "open"], ["open", "dobj", "door"], ["open", "with", "hand2"]], [["person", "verb", "put"], ["put", "dobj", "cloth"], ["put", "inside", "car"], ["put", "with", "hand1"]], [["person", "verb", "move"], ["move", "to", "cabinet"]], [["person", "verb", "open"], ["open", "dobj", "cabinet"], ["open", "with", "hand1"]], [["person", "verb", "pick-up"], ["pick-up", "with", "hand1"], ["pick-up", "dobj", "cloth"]], [["person", "verb", "move"], ["move", "to", "wall"], ["hand1", "in", "cloth"], ["move", "with", "hand1"]], [["person", "verb", "pick"], ["pick", "with", "hand2"], ["pick", "from", "wall"], ["pick", "dobj", "mop-stick"]]]\\
    \hspace{0.2cm}Question: What object was picked up before sweeping the floor?\\
    
    Example Output 10:\\
    \hspace{0.2cm}[mop-stick]\\
    
    Input:\\
    \hspace{0.2cm}Scene Graph: \{scene\_graph\}\\
    \hspace{0.2cm}Question: \{question\}\\
    
    Output:
    
\end{tcolorbox}
}
\caption{Few-shot prompt for scene graph qusetion answering tasks.}
\label{fig:Few-shot prompt for Scene Graph Qusetion Answering}
\end{figure*}
\begin{figure*}[t!]
\small
\centering
\resizebox{0.86\textwidth}{!}{
\begin{tcolorbox}[
colframe=black,        
colback=gray!10,
arc=2mm,
boxrule=1.5pt,   
title=\textbf{Prompts},
fonttitle=\bfseries
]
    You are an AI that analyzes a Scene Graph based on the context and select the best text description of it among the given candidates.\\

    1. Input:\\
    \hspace{0.2cm}- Context: A list of scene graphs representing the preceding context.\\
    \hspace{0.2cm}\hspace{0.2cm}- Each graph is composed of a set of triplets [node1, edge, node2]. `node1` and `node2` are one of person, action, object and hand. `edge` represents the relationship between them (e.g., `verb`, `dobj`, `from`, `with`).\\
    \hspace{0.2cm}- Target Scene Graph: A set of triplets that should be described into text correctly.\\
    \hspace{0.2cm}- Description Candidates: Candidates of sentence descriptions of the Target Scene Graph based on the Context.\\
    
    2. Task:\\
    \hspace{0.2cm}- Determine which description best matches the Target Scene Graph.\\
    
    3. Output:\\
    \hspace{0.2cm}- Be sure to choose only one letter of the matching description.\\
    \hspace{0.2cm}- Do not output any additional text or explanation. Only the letter in [ ] (e.g., [A]).\\
    
    Key rules of edges in a triplet:\\
    \hspace{0.2cm}- `verb` describes the action performed by `person`.\\
    \hspace{0.2cm}- `dobj` links the action to its direct object (`node2`).\\
    \hspace{0.2cm}- Other edges like `from` and `with` describe spatial relationships between nodes.\\
    
    Example Input 1:\\
    \hspace{0.2cm}- Context: [[['arrange', 'with', 'hand1'], ['arrange', 'with', 'hand2'], ['arrange', 'dobj', 'book'], ['person', 'verb', 'arrange']], [['put', 'dobj', 'book'], ['put', 'on', 'floor'], ['put', 'with', 'hand1'], ['put', 'with', 'hand2'], ['person', 'verb', 'put']], [['put', 'dobj', 'book'], ['put', 'on', 'bookshelf'], ['put', 'with', 'hand1'], ['put', 'with', 'hand2'], ['person', 'verb', 'put']]]\\
    \hspace{0.2cm}- Target Scene Graph: [['align', 'with', 'hand1'], ['align', 'with', 'hand2'], ['align', 'dobj', 'book'], ['align', 'on', 'bookshelf'], ['person', 'verb', 'align']]\\
    \hspace{0.2cm}- Description Candidates:\\
    \hspace{0.2cm}A: Finally, the flask was set down.\\
    \hspace{0.2cm}B: It was then aligned neatly on the shelf using both hands.\\
    \hspace{0.2cm}C: After achieving the desired consistency, the stick was removed from the paint can.\\
    \hspace{0.2cm}D: The cord was then sliced using the cord cutter.\\
    \hspace{0.2cm}E: Once the massaging was complete, the batter was positioned back on the counter with both hands.\\
    
    Example Output 1:\\
    \hspace{0.2cm}[B]\\
    
    ...\\
    
    Example Input 10:\\
    \hspace{0.2cm}- Context: [[['pick-up', 'with', 'hand1'], ['pick-up', 'with', 'hand2'], ['pick-up', 'dobj', 'rope'], ['person', 'verb', 'pick-up']], [['tie', 'with', 'hand1'], ['tie', 'with', 'hand2'], ['tie', 'dobj', 'rope'], ['tie', 'around', 'plant'], ['person', 'verb', 'tie']]]\\
    \hspace{0.2cm}- Target Scene Graph: [['pull', 'with', 'hand1'], ['pull', 'with', 'hand2'], ['pull', 'dobj', 'rope'], ['person', 'verb', 'pull'], ['pull', 'to', 'tighten']]\\
    \hspace{0.2cm}- Description Candidates:\\
    \hspace{0.2cm}A: The rope was then pushed loose with both hands to ensure a relaxed hold.\\
    \hspace{0.2cm}B: The rope was then pulled slack with one hand to ensure a loose grip.\\
    \hspace{0.2cm}C: The rope was then pulled tight with both hands to ensure a firm grip.\\
    \hspace{0.2cm}D: The rope was then dropped with both hands to ensure it stayed untightened.\\
    \hspace{0.2cm}E: The rope was then pulled apart with no hands to ensure it remained loose.\\
    
    Example Output 10:\\
    \hspace{0.2cm}[C]\\
    
    Input:\\
    - Context: \{context\}\\
    - Target Scene Graph: \{triplet\}\\
    - Description Candidates:\\
    \{sentences\}\\
    
    Output:
    
\end{tcolorbox}
}
\caption{Few-shot prompt for single action scene graph generation tasks.}
\label{fig:Few-shot prompt for single graph generation}
\end{figure*}
\begin{figure*}[t!]
\small
\centering
\resizebox{0.75\textwidth}{!}{
\begin{tcolorbox}[
colframe=black,        
colback=gray!10,
arc=2mm,
boxrule=1.5pt,   
title=\textbf{Prompts},
fonttitle=\bfseries
]
    You are an AI model tasked with generating scene graphs based on a given sentence. Your goal is to create exactly the specified number of scene graphs by extracting meaningful relationships between entities, actions, and objects while ensuring that the scene graphs represent actions that would visually appear in a scene.\\
    
    Rules for Generating Multiple Scene Graphs:\\
    1. Generate precisely \{num\_scene\_graphs\} scene graphs---no more, no less.\\
    2. Each scene graph must depict an action that would be explicitly visible in a scene.\\
    3. If the sentence contains multiple implicit actions, distribute them among the scene graphs while ensuring the total count matches \{num\_scene\_graphs\}.\\
    4. If there are fewer visible actions than \{num\_scene\_graphs\}, \*\*additional relevant actions may be inferred\*\* to reach the required count.\\
    5. However, use only the "Available Nodes" and "Available Edges" provided. \*\*If a necessary node is missing, use the closest semantically matching node from the available list\*\*.\\
    6. Ensure each graph maintains logical coherence while including essential contextual elements.\\
    
    Rules for A Scene Graph Representation:\\
    1. A graph is composed of one or more triplets of nodes and edges.\\
    2. A triplet starts with a node and another node is connected by an edge. (Format: node -\textgreater{} edge -\textgreater{} node)\\
    3. Each triplet is split with a new line.\\
    4. There must be exactly one triplet that starts with a person node in a graph.\\
    5. All nodes and edges must be one of "Available nodes" or "Available edges" provided.\\
    
    Rules for Node:\\
    1. A node can be person, any action, any object, or any hand.\\
    2. A node may appear explicitly or be hidden implicitly in the given sentence. Consider the context to identify the node from the "Available nodes" list, but do not create a new one.\\
    3. Map synonyms or semantically similar words to nodes in the "Available nodes" list.\\
    4. Use default tools or body parts for actions that imply them (e.g., hands for grasping).\\
    5. Treat each action as a node.\\
    6. Include "person" as the starting node in the graph.\\
    
    Special Rules for Hand Node:\\
    1. If both hands are empty and a node is grasped, represent it as "hand1."\\
    2. If one hand holds a node and another node is grasped, represent it as "hand2."\\
    3. If all hands release their objects, reset the next grasping hand to "hand1."\\
    4. Ensure "hand1" and "hand2" are used contextually to avoid overlap or ambiguity.\\
    5. If the sentence implies using both hands (e.g., lifting a large object), represent both hands explicitly (e.g., hand1, hand2).\\
    
    Rules for Edge:\\
    1. An edge can be verb, dobj, or any preposition.\\
    2. Use only the edges listed under "Available edges."\\
    3. Here are the explanations for each edge.\\
    \hspace{0.2cm}- verb: can only connect person and an action node. (e.g., person -\textgreater{} verb -\textgreater{} add)\\
    \hspace{0.2cm}- dobj: connects an action and an object node, only when it is the direct object of the action (e.g., add -\textgreater{} dobj -\textgreater{} flour)\\
    \hspace{0.2cm}- preposition: connects one of the four types of node pairs: action \& object / action \& hand / object \& object / hand \& object (e.g., take -\textgreater{} from -\textgreater{} table)\\
    
    Output Format:\\
    The output must consist of exactly \{num\_scene\_graphs\} scene graphs, each separated with a blank line. For a graph, output one triplet per line. Follow the format below (an example of three scene graphs of multiple triplets):\\
    node -\textgreater{} edge -\textgreater{} node\\
    node -\textgreater{} edge -\textgreater{} node\\
    ...\\
    
    node -\textgreater{} edge -\textgreater{} node\\
    node -\textgreater{} edge -\textgreater{} node\\
    ...\\
    
    node -\textgreater{} edge -\textgreater{} node\\
    node -\textgreater{} edge -\textgreater{} node\\
    ...\\
    
    Use only the "Available Nodes" and "Available Edges" provided. No additional text, explanations, or formatting should be included.\\
    
    Example Input 1:\\
    ...\\
    
    Example Input 10:\\
    \hspace{0.2cm}- Context: The hole was carefully aligned to ensure a secure fit. A screwdriver was selected from the toolbox and positioned against the screw head.\\
    \hspace{0.2cm}- Target sentence: With a steady hand, the screwdriver was twisted, driving the screw into place while the piece was firmly held.\\
    \hspace{0.2cm}- Available nodes: person, screw, wood, hand1, hand2, hole, screwdriver, toolbox, align, select, position, hold, twist\\
    \hspace{0.2cm}- Available edges: verb, dobj, into, with, from, against\\
    \hspace{0.2cm}- Number of Scene Graphs: 2\\
    
    Example Output 10:\\
    person -\textgreater{} verb -\textgreater{} twist\\
    twist -\textgreater{} dobj -\textgreater{} screwdriver\\
    twist -\textgreater{} with -\textgreater{} hand2\\
    
    Input:\\
    \hspace{0.2cm}- Context: \{context\}\\
    \hspace{0.2cm}- Target sentence: \{target\_sentence\}\\
    \hspace{0.2cm}- Available nodes: \{available\_nodes\}\\
    \hspace{0.2cm}- Available edges: \{available\_edges\}\\
    \hspace{0.2cm}- Number of Scene Graphs: \{num\_scene\_graphs\}\\
    
    Output:
    
\end{tcolorbox}
}
\caption{Few-shot prompt for multiple action scene graph generation tasks.}
\label{fig:Few-shot prompt for multi graph generation}
\end{figure*}

\end{document}